\pdfoutput=1
\documentclass{article}
\usepackage{iclr2027_conference,times}

\usepackage{amsmath,amsfonts,bm}

\def\eqref#1{equation~\ref{#1}}

\def\1{\bm{1}}

\DeclareMathAlphabet{\mathsfit}{\encodingdefault}{\sfdefault}{m}{sl}
\SetMathAlphabet{\mathsfit}{bold}{\encodingdefault}{\sfdefault}{bx}{n}

\newcommand{\E}{\mathbb{E}}

\newcommand{\R}{\mathbb{R}}

\newcommand{\softmax}{\mathrm{softmax}}

\newcommand{\Cov}{\mathrm{Cov}}

\usepackage{hyperref}
\hypersetup{hidelinks}
\usepackage{url}
\usepackage{amsmath,amssymb,amsthm,mathtools}
\usepackage{booktabs}
\usepackage{multirow}
\usepackage{graphicx}
\usepackage{xcolor}
\usepackage[section]{placeins}
\usepackage{microtype}
\usepackage{flafter}
\usepackage{enumitem}

\newtheorem{theorem}{Theorem}
\newtheorem{proposition}[theorem]{Proposition}
\newtheorem{corollary}[theorem]{Corollary}
\newtheorem{lemma}[theorem]{Lemma}
\theoremstyle{definition}

\theoremstyle{remark}
\newtheorem{remark}[theorem]{Remark}

\newcommand{\Prob}{\mathsf{Pr}}
\newcommand{\PH}{\mathrm{PH}}
\newcommand{\base}{\mathrm{base}}
\newcommand{\corr}{\mathrm{corr}}
\newcommand{\apply}{\mathrm{apply}}

\title{Risk-Controlled Correction for Frozen\\ Multimodal Models under Missing Modalities}

\author{Hoang Pham Cong$^{1}$, Long D. Nguyen$^{2}$, Ho Viet Duc Luong$^{1}$, \\
\textbf{Nguyen Tran Minh Nhat$^{1}$, Binh P. Nguyen$^{2}$ \& Huynh Thi Thanh Binh$^{1}$} \\
$^{1}$Hanoi University of Science and Technology, Hanoi, Vietnam \\
$^{2}$Victoria University of Wellington, Wellington, New Zealand}
\hypersetup{pdftitle={Risk-Controlled Correction for Frozen Multimodal Models under Missing Modalities},
pdfauthor={Hoang Pham Cong, Long D. Nguyen, Ho Viet Duc Luong, Nguyen Tran Minh Nhat, Binh P. Nguyen, Huynh Thi Thanh Binh}}

\iclrfinalcopy
\begin{document}
\addtocontents{toc}{\protect\setcounter{tocdepth}{-1}}
\maketitle
\lhead{Preprint}

\begin{abstract}

A deployed multimodal predictor may face missing modalities after its weights can no longer be changed. This raises a question that training-time repair does not answer: \emph{what part of the resulting error is still correctable from the information that remains?} We show that this part is characterized by the \emph{Bayes gap}, the difference between the model's prediction and the Bayes prediction given the observed modalities. For cross-entropy and squared error, excess risk measures this gap in the geometry of the loss; ambiguity in the observed modalities cannot be removed. Thus, a badly degraded prediction is not necessarily a correctable one. Estimating the Bayes prediction normally requires labels. However, when a model observing more modalities is conditionally correct, averaging its outputs conditional on the remaining modalities recovers this target without retrieval labels; with the same fixed blend and neighbors, it is no worse in expected cross-entropy or Brier risk than averaging hard labels. Some such assumption is unavoidable: otherwise the target is not identifiable from unlabeled data alone. GUARD (Gated Utility-Aware Recalibration for Deployment) leaves the deployed predictor unchanged. It estimates a candidate correction by retrieval, learns from deployment-matched labeled data when correction is likely to help, and calibrates a threshold on a held-out split by conformal risk control. Under exchangeability, the marginal probability that GUARD applies a correction whose loss increase exceeds a prescribed tolerance is bounded by a preset level, regardless of the learned score. In controlled experiments, correction gain follows headroom rather than raw error; across nine frozen model--dataset pairs spanning six domains, GUARD retains most of blanket correction's gain within harm budgets that blanket correction often exceeds.

\end{abstract}

\section{Introduction}
\label{sec:intro}

\looseness=-1 Modalities can be missing at deployment. A diagnostic model trained on twelve-lead ECG~\citep{wagner2020ptbxl} may run where only a few leads are recorded~\citep{reyna2021cinc}; an activity recognizer trained on a body-worn sensor array~\citep{ordonez2016deepconvlstm} may lose streams when sensors disconnect; an audio--visual model may receive only one of its two streams~\citep{parthasarathy2020missing}. Deployments may also fix the predictor because its weights are inaccessible~\citep{sun2022blackbox}. We assume query access to predictions and representations, without modifying the model. Output correction still requires validation and applicable change control~\citep{fda2025pccp}.

\begin{figure}[!htbp]
\centering
\includegraphics[width=\linewidth]{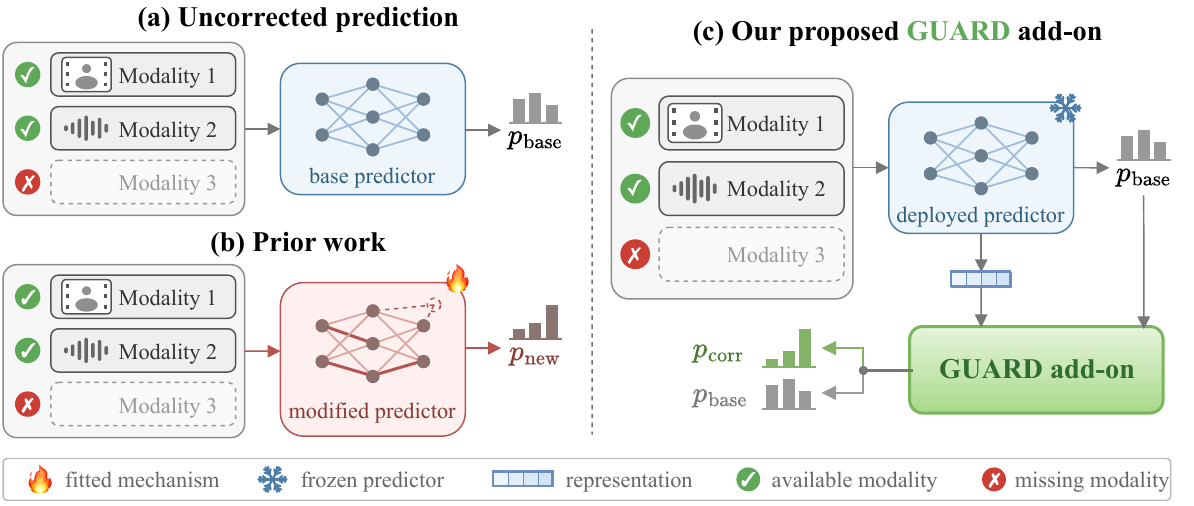}
\caption{\textbf{Missing-modality prediction and post-hoc correction.} \emph{(a)} Uncorrected prediction under missing modalities. \emph{(b)} Prior methods insert a fitted mechanism into the prediction pipeline before deployment. \emph{(c)} GUARD leaves the deployed predictor unchanged and selectively corrects its output.}
\label{fig:overview}
\end{figure}

\looseness=-1 Existing missing-modality methods reconstruct absent modalities~\citep{wang2023imder}, distill complete-modality teachers~\citep{li2024corrkd,zhuang2025cmad}, align partial and complete representations~\citep{lin2023missmodal,he2026ease}, retrieve full-modality neighbors~\citep{zhuang2025hme,zhi2025icl}, or learn reliability-aware routing~\citep{zou2026dear}. Recent work also asks whether repair is needed at all~\citep{chen2026ebg,madaan2026primo,gao2026sieve}. These approaches generally modify or augment the prediction pipeline before deployment, even when the backbone remains frozen (Figure~\ref{fig:overview}b). We instead study post-hoc output correction that leaves the deployed predictor unchanged (Figure~\ref{fig:overview}c). This raises a deployment question: \emph{once the model is frozen, what part of its error is still correctable from the modalities it observes?}

\looseness=-1 This paper studies a deployment regime in which the deployed predictor must remain fixed, but labeled data from the target deployment condition are available for post-hoc adaptation. We use these data to fit the correction mechanism and calibrate without modifying or retraining the predictor.

\looseness=-1 The first distinction is between \emph{error} and \emph{correctable error}. Missingness can expose systematic bias, but it can also create ambiguity: different outcomes may be compatible with the same observed modalities. Only the former is reachable by a correction computed from what remains. We show that this reachable component is exactly the \emph{Bayes gap}, the discrepancy between the frozen prediction and the best prediction available from the observed modalities. Consequently, a badly degraded prediction need not offer more correction headroom than a mildly degraded one.

\looseness=-1 The second distinction concerns where a correction target can come from. For nested observations, we derive an exact relation between their Bayes gaps. When a predictor with more modalities is conditionally correct, its outputs can replace retrieval labels: averaging richer-view outputs conditional on the remaining modalities recovers the poorer-view Bayes target and, for the same fixed blend and neighbors, is no worse in expected cross-entropy or Brier risk than averaging hard labels. This assumption cannot be dropped: without some link between richer outputs and labels, the poorer Bayes target is not identifiable from unlabeled paired data alone.

\looseness=-1 The third distinction is between \emph{correction} and \emph{intervention}. A candidate correction that helps on average can still harm individual predictions. Our method, \textbf{GUARD}, therefore separates the two decisions: \textsc{Recalibrate} constructs a candidate output correction, while \textsc{Certify} decides whether that candidate may replace the frozen prediction. GUARD learns an action score from deployment-available quantities and calibrates a threshold on a separate held-out split by conformal risk control~\citep{angelopoulos2024crc}. Under exchangeability, the marginal probability that GUARD both intervenes and increases loss beyond a prescribed tolerance is bounded by a preset level, for any score fixed before calibration.

Our contributions are as follows.
\begin{itemize}[leftmargin=1.4em,itemsep=1pt,topsep=2pt]
    \item \textbf{Correctability.} For common canonical losses, we characterize the error that remains correctable after a multimodal predictor is frozen: deployable correction can act only on the Bayes gap, not on ambiguity induced by the observed modalities.
    \item \textbf{Cross-mask transfer.} We derive an exact relation between the Bayes gaps of nested modality sets. It yields a retrieval target from richer-view frozen outputs, establishes when it exactly recovers the poorer-view Bayes predictor, proves finite-sample dominance for matched fixed blends under a conditionally correct richer predictor, and explains why retrieval-label-free recovery needs an assumption.
    \item \textbf{Risk-controlled intervention.} GUARD separates constructing a correction from deciding whether to use it. Its action score ranks utility, while an independent conformal calibration controls the marginal probability of harmful intervention for any fixed scorer and corrector.
    \item \textbf{Empirical evaluation.} Controlled experiments show that gain follows correctable headroom rather than raw error, and across nine frozen model--dataset pairs in six domains GUARD keeps most of blanket correction's utility while controlling harmful intervention.
\end{itemize}

\section{Related Work}
\label{sec:related}

\paragraph{Missing-modality learning.}

Methods for handling absent modalities include aligning representations~\citep{lin2023missmodal}, distilling from a complete-modality teacher~\citep{li2024corrkd,zhuang2025cmad}, completing inputs by prompting~\citep{lee2023mapmm,guo2024mplmm}, retrieving across samples~\citep{zhuang2025hme}, and building reliability-aware architectures~\citep{he2026ease,zou2026dear,zhang2024lnln,zhuang2026tmdc}. \citet{ma2022robust} document how sharply multimodal transformers degrade when a modality is withheld. A recent line asks instead whether a missing modality should be repaired at all~\citep{chen2026ebg,madaan2026primo,gao2026sieve}.
Closest in motivation, \citet{gao2026sieve} learn a gate jointly with paired direct and repaired branches but provide no finite-sample guarantee on harmful interventions. GUARD addresses a related question under a different deployment constraint: the deployed weights remain unchanged, the gate is fitted post hoc and calibrated on separate data, and no modality is reconstructed.

\paragraph{Correction after a frozen predictor.}

\citet{roth2026fedcore} also freeze the deployed predictor, but fit residual completion operators from paired complete and incomplete views. GUARD instead edits outputs directly and couples correction with finite-sample harm control; its default hard-label retrieval target does not require paired complete and incomplete views. Other ways to adapt a deployed system change the predictor or its outputs rather than deciding when to act: test-time adaptation on unlabeled target data~\citep{wang2021tent,ramazanova2025midl}, label-shift correction~\citep{lipton2018labelshift}, and nearest-neighbor interpolation such as $k$NN-LM~\citep{khandelwal2020knnlm}, which motivates our default retrieval implementation. Selective recalibration~\citep{zollo2024selective} does learn which inputs to recalibrate, but it targets calibration error on the selected inputs rather than a bound on harmful intervention.

\paragraph{Selective intervention and risk control.}

Deciding whether to act connects to selective prediction, where rejection is learned jointly with prediction~\citep{geifman2019selectivenet} or samples are deferred to another decision maker~\citep{madras2018defer}, and to distribution-free risk control~\citep{angelopoulos2021ltt,angelopoulos2024crc}, with loss-controlling and action-conditional extensions~\citep{wang2023clcp,laufer2023pareto,zhu2026action}. Conformal decision theory is closest in form, calibrating decisions directly against a risk budget rather than through prediction sets~\citep{lekeufack2024cdt}; \citet{moayedikia2026conformal} instead certifies the prediction sets of a multimodal fusion architecture. GUARD scores each candidate intervention on a frozen predictor and calibrates one threshold for it, which yields the finite-sample harm guarantee stated in \S\ref{sec:certify}.

\section{Correctability After the Model is Frozen}
\label{sec:reduction}
\subsection{Frozen deployment setting}

A sample carries $M$ modalities and a label $Y$. At deployment a mask $S\subseteq\{1,\ldots,M\}$ indicates the observed ones, and $X=(x_S,S)$ collects what is observed. Everything the corrector may read at deployment, including any representation the frozen model computes, is a function of $X$. A frozen model turns $X$ into scores $z\in\R^K$, where $K$ is the output dimension ($K=C$ for $C$-class classification).
In this paper, we consider \emph{canonical} losses $\ell(y,z)$ whose score-space gradient has the form
\begin{equation}
\label{eq:canonical}
    \nabla_z\ell(y,z)=m(z)-t(y),
\end{equation}
where $m(\cdot)$ is the prediction and $t(\cdot)$ is the label statistic. For example, cross-entropy loss with softmax satisfies $m(z)=\softmax(z)$, $t(y)=e_y$; half-squared error $\ell(y,z)=\frac12\|z-y\|^2$ satisfies $m(z)=z$, $t(y)=y$. We write $\bar\ell(y,p)$ for the loss evaluated on predictions, e.g.\ $-\log p_y$ for cross-entropy; loss conventions are detailed in App.~\ref{app:loss-conventions}.

\subsection{The Bayes gap: correctable error is not raw error}

\label{sec:principle1}
Condition \eqref{eq:canonical} makes the deployment question tractable: averaging over labels replaces $t(Y)$ by its conditional mean. Define
\begin{equation}
\label{eq:pid}
    \pi(X):=\E[t(Y)\mid X],\qquad d(X):=m(z(X))-\pi(X).
\end{equation}
Here $\pi(X)$ is the conditional class distribution for classification or conditional mean for squared error, and $d(X)$ is the model's \emph{Bayes gap} from that target.

\begin{theorem}[Bayes-gap reduction]
\label{thm:reduction}
Let $\ell(y,\cdot)$ be differentiable and convex on $\R^K$, satisfy \eqref{eq:canonical}, and have finite conditional risk and integrable $t(Y)$. For any finite deployable output $z'(X)$,
\begin{equation}
\label{eq:exact-principle}
    \nabla_{z'}\,\E[\ell(Y,z')\mid X]
    =\E[m(z')-t(Y)\mid X]
    =m(z')-\pi(X).
\end{equation}
Hence every finite conditional optimum satisfies $m(z^\star(X))=\pi(X)$, and every such stationary point is optimal. An $X$-measurable post-hoc correction can therefore remove only the excess risk carried by the Bayes gap $d(X)$. In particular, if $m(z)=\pi(X)$ almost surely, no deployable output correction based on $X$ can reduce expected risk. Boundary class distributions may require limiting logits.
\end{theorem}

The proof is in App.~\ref{app:proof-reduction}. The underlying geometry is classical~\citep{banerjee2005bregman,gneiting2007proper}; the deployment consequence is not. The result is exact in output space: uncertainty remaining at $\pi(X)$ cannot be eliminated by a correction based on $X$. Representation-level corrections face the same information constraint and induce measurable output edits (Props.~\ref{prop:feature},~\ref{prop:output}). Thus, \emph{degradation severity and correction headroom are different quantities}: a severely degraded predictor may have little systematic error left to remove, while a less degraded one may retain substantial bias. \S\ref{sec:prediction} examines this distinction.

\subsection{Nested observations: when richer views can replace labels}
\label{sec:inherit}

Theorem~\ref{thm:reduction} identifies $\pi(X)$, written $\pi_S$ under an observed modality set $S$, as the target of any correction, but estimating $\pi_S$ ordinarily requires
labels. Multimodal data provide an additional source of information: the same
sample may also be observed under a richer modality set $T\supseteq S$.
This raises a natural question: \emph{can predictions made from the richer view
serve as targets for correcting the poorer one?}
Write $X_S$, $m_S$, $\pi_S$, and $d_S=m_S-\pi_S$ for the quantities associated
with the poorer mask $S$, and similarly for $T$. Assume $\sigma(X_S)\subseteq\sigma(X_T)$ under the same joint law, including any mask context. The tower property gives
$\pi_S=\E[\pi_T\mid X_S]$ and hence
\begin{equation}
\label{eq:inherit}
    d_S
    =
    \underbrace{\bigl(m_S-\E[m_T\mid X_S]\bigr)}
    _{\text{estimable from paired views and outputs}}
    +
    \underbrace{\E[d_T\mid X_S]}
    _{\text{inherited richer-view error}} .
\end{equation}
The first term can be estimated from paired views and frozen predictions, whereas
the second measures the systematic error inherited from the richer-view
predictor. If the richer-view error vanishes after conditioning on $X_S$,
that is, $\E[d_T\mid X_S]=0$, then
$ \pi_S=\E[m_T\mid X_S]$.
Thus, the population conditional average recovers the Bayes target for $S$ without retrieval labels; finite-neighbor estimates additionally incur estimation error. A sufficient special case is a richer-mask predictor that is itself Bayes-optimal, $m_T=\pi_T$; that stronger condition is also the one the dominance result of \S\ref{sec:correct} uses. Some assumption of this kind is unavoidable: without one, $\pi_S$ is not identifiable from the unlabeled paired law (App.~\ref{app:inherit}). Neither identity says whether a given correction should be applied.

\section{GUARD: From Correction to Intervention}
\label{sec:guard}

\begin{figure}[!htbp]
\centering
\includegraphics[width=\linewidth]{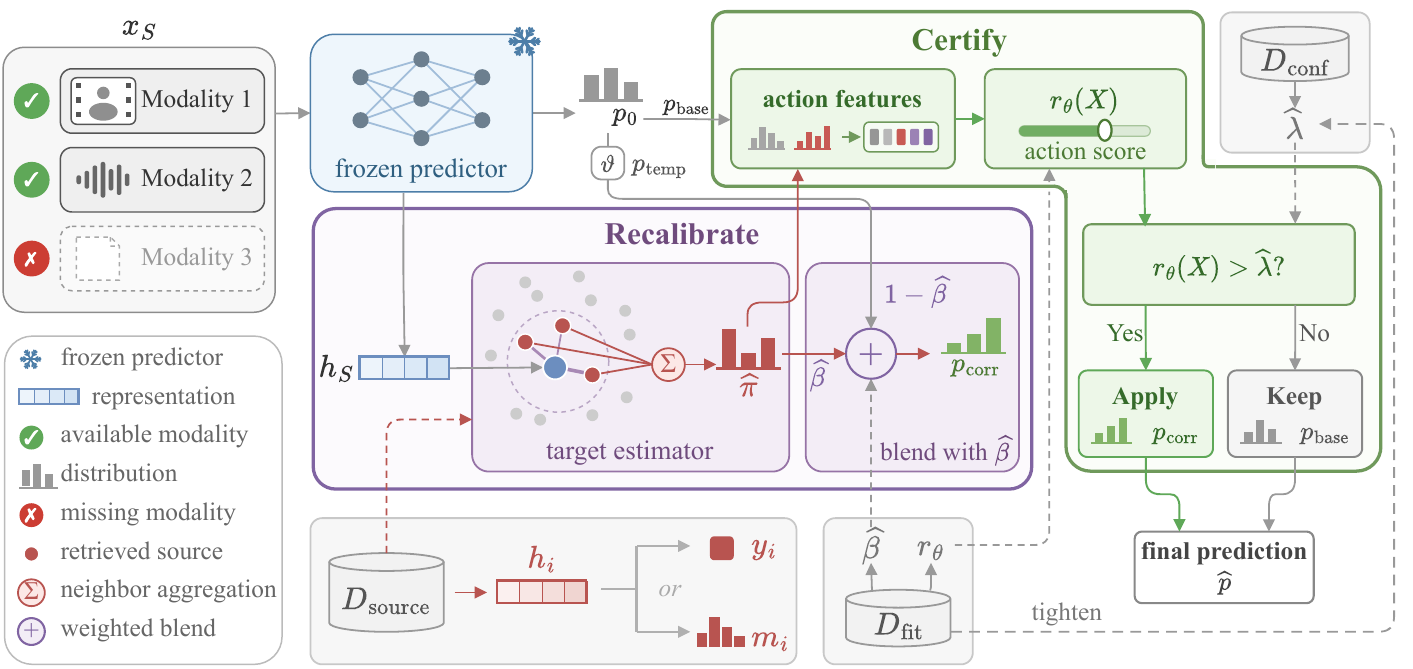}
\caption{\textbf{GUARD at deployment.} The frozen output $p_0$ is temperature-scaled to $p_{\mathrm{temp}}$ before blending with $\widehat\pi$. The reference $p_{\base}=p_0$ feeds the action features, defines harm on $D_{\mathrm{conf}}$, and is kept when the gate declines (App.~\ref{app:reference}). Offline, $D_{\mathrm{source}}$ builds the target estimator; $D_{\mathrm{fit}}$ selects $\vartheta$ and $\widehat\beta$, fits $r_\theta$, and tightens the threshold that $D_{\mathrm{conf}}$ calibrates by conformal risk control.}
\label{fig:guard}
\end{figure}

\looseness=-1 The theory answers what can in principle be corrected, but deployment still requires two separate decisions: how to construct a candidate correction, and whether that candidate should replace the frozen prediction. GUARD separates them: without touching model weights, \textsc{Recalibrate} turns the Bayes target into an estimable candidate prediction, while \textsc{Certify} treats that candidate as an optional action and decides when it may be applied under a prescribed harm budget. Three disjoint splits carry the offline steps: $D_{\mathrm{source}}$ builds the target estimator, $D_{\mathrm{fit}}$ fits the blend weight $\widehat\beta$ and action score $r_\theta$, and $D_{\mathrm{conf}}$ calibrates the threshold $\widehat\lambda$. Figure~\ref{fig:guard} illustrates the implementation of GUARD.
\subsection{Recalibrate: construct a candidate correction}
\label{sec:recal}
\label{sec:correct}

\looseness=-1 Given the Bayes target $\pi(X)$ identified in \S\ref{sec:reduction}, GUARD treats its estimator as a frozen plug-in
$\widehat\pi=\mathcal A(D_{\mathrm{source}},X)$ built from a source split. We instantiate
$\mathcal A$ by averaging neighbors in the available feature space; its information limit is discussed in App.~\ref{sec:limitations}. The default target averages their
labels, while the inheritance identity \eqref{eq:inherit} yields a second target that replaces labels
with predictions from a richer modality set:
\begin{equation}
\label{eq:knn2}
    \text{(hard-label)}\ \widehat\pi^{(Y)}(X)=\!\!\sum_{j\in\mathcal N_k(X)}\!\!w_j\,t(Y_j),
    \hspace{0.8em}
    \text{(cross-mask transfer)}\ \widehat\pi^{(T)}(X)=\!\!\sum_{j\in\mathcal N_k(X)}\!\!w_j\,m_T\!\bigl(X_T^{(j)}\bigr),
\end{equation}
where $w_j\ge0$ and $\sum_jw_j=1$. Let $p_{\mathrm{temp}}=\operatorname{Temp}_{\vartheta}(m(z))$ denote the frozen prediction after temperature scaling. The weights, temperature $\vartheta$, $k$, target and retrieval space are selected on $D_{\mathrm{fit}}$ (App.~\ref{app:setup}). Either target produces a candidate
\begin{equation}
\label{eq:blend}
    p_{\corr}=(1-\widehat\beta)\,p_{\mathrm{temp}}+\widehat\beta\,\widehat\pi,
\end{equation}
where $\widehat\beta\in[0,1]$ minimizes mean prediction loss $\bar\ell$ on $D_{\mathrm{fit}}$ (App.~\ref{app:erm}). The reference $p_{\base}$ used to measure harm and retained on rejected queries is the frozen output $p_0$ throughout (App.~\ref{app:reference}). Selecting the candidate affects utility but not the validity of \textsc{Certify} under its exchangeability assumption. The hard-label mode is nearest-neighbor interpolation in the spirit of $k$NN-LM~\citep{khandelwal2020knnlm}; App.~\ref{sec:corrector} also evaluates a trained probe.

\paragraph{Why richer-mask outputs can replace labels.}
The two targets exchange different sources of error. Hard-label retrieval contains per-neighbor
label noise; cross-mask transfer removes that noise but inherits any systematic error of the
richer-view predictor. If that predictor is conditionally correct, $m_T=\pi_T$, its output is the
conditional mean of the corresponding hard label. Prop.~\ref{prop:transfer} then shows that,
under cross-entropy or Brier score, the cross-mask blend has no larger expected risk for the same fixed $\beta\in[0,1]$, admissible $k$, neighbors and feature-based weights. This compares fixed blends before gating; it does not order separately tuned or tempered targets. App.~\ref{app:transfer} gives the proof, optimal blend, and
label-efficiency trade-off.

\subsection{Certify: deciding when to apply it}
\label{sec:certify}

\paragraph{From correction to intervention.}
A correction that improves average risk may still harm individual predictions, so GUARD treats
$p_{\corr}$ as a candidate action rather than an automatic replacement. For a labeled pair
$(X,Y)$, define the change in loss induced by correction as
\begin{equation}
\label{eq:deltay}
    \Delta_Y(X)=\bar\ell(Y,p_{\corr}(X))-\bar\ell(Y,p_{\base}(X)).
\end{equation}
Thus $\Delta_Y(X)<0$ means that correction helps, while we call an intervention
\emph{harmful} when $\Delta_Y(X)>\delta$, for a tolerated increase $\delta\ge0$.
An ideal policy would intervene only where correction helps, but $\Delta_Y$ depends on $Y$,
which is unavailable at deployment.
We therefore use $D_{\mathrm{fit}}$ to learn a score $r_\theta(X)$ that ranks candidate
interventions by their predicted utility. The score is trained with the binary target
\begin{equation}
Z=\mathbf 1\{\bar\ell(Y,p_{\corr}(X))<\bar\ell(Y,p_{\base}(X))\},
\end{equation}
but uses only features available at deployment. In our implementation, $r_\theta$ is a logistic
model over five features derived from $(p_{\base},\widehat\pi)$, listed in
App.~\ref{app:erm}. The score $r_\theta$ only ranks candidates: it is trained to
predict whether correction improves loss ($\Delta_Y<0$), whereas the guarantee below controls
the distinct event of excessive harm ($\Delta_Y>\delta$).

\paragraph{Finite-sample harm control.}
By itself, this ranking provides no control over
how often harmful corrections are applied. We therefore threshold it to obtain a nested family
of intervention policies,
\begin{equation}
\label{eq:policy}
    A_\lambda(X)=\mathbf 1\{r_\theta(X)>\lambda\},\qquad
    L_\lambda(X,Y)
    =A_\lambda(X)\,\mathbf 1\{\Delta_Y(X)>\delta\},
\end{equation}
where $A_\lambda$ determines whether to intervene and $L_\lambda$ indicates a harmful
intervention. As $\lambda$ increases, the intervention set can only shrink, so
$L_\lambda\in[0,1]$ is non-increasing in $\lambda$.
We calibrate $\lambda$ on a held-out split
$D_{\mathrm{conf}}=\{(X_i,Y_i)\}_{i=1}^n$, which is not used to fit the corrector,
$\widehat\beta$, or $r_\theta$. Let
$\widehat R_n(\lambda)=n^{-1}\sum_i L_\lambda(X_i,Y_i)$ be its empirical harmful-intervention
rate, and let $\alpha\in(0,1)$ be the prescribed harm budget. Conformal risk
control~\citep{angelopoulos2024crc} selects
\begin{equation}
\label{eq:crc}
    \widehat\lambda
    =\inf\Bigl\{\lambda:
    \frac{n}{n+1}\widehat R_n(\lambda)+\frac{1}{n+1}\le\alpha
    \Bigr\},
\end{equation}
with $\widehat\lambda=+\infty$ (no intervention) if the set is empty or $n=0$. The selected threshold has the following guarantee:

\begin{proposition}[Finite-sample harm control, after \citealp{angelopoulos2024crc}]
\label{thm:noharm}
Condition on all pre-calibration fitting, fixing $p_{\base}$, $p_{\corr}$, $r_\theta$, the loss and $\delta$. If calibration and deployment samples are exchangeable under this conditional law, then $\widehat\lambda$ from \eqref{eq:crc} satisfies
\begin{equation}
\label{eq:noharm}
    \Prob\!\left(\Delta_Y(X)>\delta\ \wedge\ \apply\right)\le\alpha .
\end{equation}
\end{proposition}

App.~\ref{app:noharm} verifies the conditions required by conformal risk control.
Proposition~\ref{thm:noharm} has two consequences. First, the guarantee holds for
\emph{any} score fixed before $D_{\mathrm{conf}}$ is read: a poor ranking may reduce utility
or intervention rate, but it does not invalidate the harm bound. Second, the guarantee is
marginal over the calibration draw and deployment sample. It therefore does not certify any
individual correction and controls joint harm,
$\Prob(\Delta_Y>\delta\wedge\apply)$, rather than the conditional rate
$\Prob(\Delta_Y>\delta\mid\apply)$. For corrections that exceed the tolerance, label ambiguity puts a floor under conditional harm (Prop.~\ref{prop:floor}); App.~\ref{app:conditional-harm} reports all three rates.

GUARD may further increase $\widehat\lambda$ using $D_{\mathrm{fit}}$ to remove low-ranked
interventions that add little estimated utility (App.~\ref{app:erm}). This cannot weaken the
guarantee: for any $\lambda^\star\ge\widehat\lambda$,
$L_{\lambda^\star}\le L_{\widehat\lambda}$ pointwise, and hence
$\E[L_{\lambda^\star}]\le\alpha$. The resulting certified policy applies
$p_{\corr}$ only when admitted by the gate and otherwise retains $p_{\base}$; its
\emph{gated gain} is the corresponding change in the benchmark metric.

\section{Experimental Results}
\label{sec:experiments}

\subsection{Experimental setup}
\label{sec:setup}

Our evaluation spans seven benchmark families across six application domains, covering nine
frozen model--dataset pairs and forty-four degraded conditions. Each frozen model is anchored to its published implementation or protocol before GUARD is attached, and its
parameters remain fixed throughout evaluation. The degradation axes reflect different forms of
information loss covered by Theorem~\ref{thm:reduction}: entire modalities are withheld on
multimodal benchmarks, sensor families, electrodes, or ECG leads are removed, and protein sequences are progressively truncated.

For every condition, $D_{\mathrm{source}}$, $D_{\mathrm{fit}}$, $D_{\mathrm{conf}}$, and
$D_{\mathrm{test}}$ have disjoint roles: IEMOCAP is split by video, OPPORTUNITY by time block and the others by row within the evaluated population, so exchangeability is claimed for held-out rows or blocks rather than an unseen subject (App.~\ref{app:setup}). All retrieval choices and $\widehat\beta$ are selected before test evaluation,
and \textsc{Certify} is calibrated only on $D_{\mathrm{conf}}$; the test set is read once after
all decisions are fixed. Unless stated otherwise, we use harm budget $\alpha=0.2$ and tolerance
$\delta=0.05$ and report averages over the prescribed seeds or deployment splits. The four subsections below test one claim each. App.~\ref{app:setup} gives model provenance, degradation levels,
data splits, retrieval grids, and baselines; App.~\ref{app:percond} reports all per-condition results. Code is available at \url{https://github.com/kongwoang/GUARD}.

\subsection{Does gain follow correctable headroom rather than raw error?}
\label{sec:prediction}

\begin{figure}[!htbp]
\centering
\includegraphics[width=\linewidth]{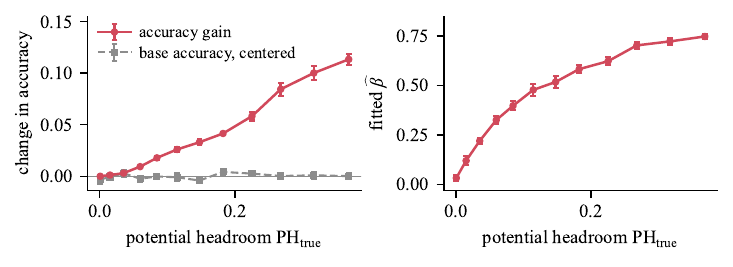}
\caption{\textbf{Gain follows the gap, not the error} on 12 synthetic configurations and 10 seeds.
For each ambiguity level $\tau$, the bias parameter $\gamma$ is chosen to keep base accuracy near
$0.65$, so the amount of error stays fixed.
\emph{Left:} accuracy gain against true potential headroom
$\PH_{\mathrm{true}}=\E\|d(X)\|_2^2$, with base accuracy centered on its mean.
\emph{Right:} the blend weight selected on $D_{\mathrm{fit}}$.}
\label{fig:synthetic}
\end{figure}

Theorem~\ref{thm:reduction} identifies the target of correction, while realized gain also depends on its estimation. To isolate headroom from raw error, we need a known Bayes target and separate control over ambiguity and bias.
We therefore construct a controlled synthetic classification with
$\pi(X)=\softmax(WX/\tau)$ and frozen-model prediction
$m(X)=\softmax(WX/\tau+\gamma VX)$. Here $\tau$ controls conditional ambiguity, while
$\gamma$ controls systematic model bias. For each $\tau$, we choose $\gamma$ so that base accuracy
remains near $0.65$, producing 12 configurations with similar raw error but different
correctable headroom.
Because $\pi(X)$ is known in this setting, we can measure the true potential headroom
$\PH_{\mathrm{true}}:=\E\|d(X)\|_2^2$, the reducible component of the unhalved Brier risk. Accuracy gain need not be monotone in this quantity; the relation below is empirical.
Across 10 seeds, correction gain tracks $\PH_{\mathrm{true}}$ with correlation $0.99$ and
vanishes as the gap disappears; the fitted blend weight decreases with it. At matched base accuracy the most bias-dominated configuration gains $+0.11$ in accuracy and $+0.86$ in loss, while the ambiguity-dominated one gains nothing on either scale.

\subsection{Does post-hoc correction generalize across frozen models?}\label{sec:crossmodels}

We next ask whether the same frozen-deployment policy survives changes in architecture, modality type and application domain. On CMU-MOSEI, attaching GUARD to three published frozen models improves all three, by up to $2.1$ accuracy points on average; the quoted rows are context, not a ranking: they use their own protocols and no deployment-matched labels. The gain concentrates on the subsets that lose language, averaging $3.5$ points there and reaching $5.8$, and stays within a point when language is retained. Table~\ref{tab:mosei-main} gives the full published comparison; the main question here is whether the same behavior extends beyond sentiment.

\begin{table}[!htbp]
\caption{\textbf{CMU-MOSEI under different modality combinations.} Binary accuracy and weighted F1 (\textsc{acc}/F1, \%) on non-neutral samples. $L$, $A$ and $V$ denote observed language, acoustic and visual modalities. Unmarked rows are quoted from \citet{zhuang2025cmad} and \citet{zhuang2026tmdc}; $\dagger$ marks our frozen reproductions. Reproduced rows average five deployment splits, except the unchanged text-only CMAD rows, which use the full test set. Averages are taken before rounding; bold marks the best entry in each column. Harm and apply rates are in App.~\ref{app:percond}.}
\label{tab:mosei-main}
\centering\scriptsize
\setlength{\tabcolsep}{2.4pt}
\begin{tabular}{@{}lccccccc c@{}}
\toprule
\textbf{Method} & $L$ & $A$ & $V$ & $L,A$ & $L,V$ & $A,V$ & $L,A,V$ & \textbf{Avg.} \\
\midrule
\multicolumn{9}{@{}l}{\emph{Published under their own training protocols; context, not a controlled comparison}}\\
MCTN~\citep{pham2019mctn}                & 82.6/82.8 & 62.7/54.5 & 62.6/57.1 & 83.5/83.3 & 83.2/83.2 & 63.7/62.7 & 84.2/84.2 & 74.6/72.5 \\
MMIN~\citep{zhao2021mmin}                & 82.3/82.4 & 58.9/59.5 & 59.3/60.0 & 83.7/83.3 & 83.8/83.4 & 63.5/61.9 & 84.3/84.2 & 73.7/73.5 \\
GCNet~\citep{lian2023gcnet}              & 83.0/83.2 & 60.2/60.3 & 61.9/61.6 & 84.3/84.4 & 84.3/84.4 & 64.1/57.2 & 85.2/85.1 & 74.7/73.7 \\
IMDer~\citep{wang2023imder}              & 84.5/84.5 & 63.8/60.6 & 63.9/63.6 & 85.1/85.1 & 85.0/85.0 & 64.9/63.5 & 85.1/85.1 & 76.0/75.3 \\
DiCMoR~\citep{wang2023dicmor}            & 84.2/84.3 & 62.9/60.4 & 63.6/63.6 & 85.0/84.9 & 84.9/84.9 & 65.2/64.4 & 85.1/85.1 & 75.8/75.4 \\
MPLMM~\citep{guo2024mplmm}               & 84.6/84.6 & 61.8/49.4 & 62.2/53.2 & 85.1/85.1 & 85.2/85.1 & 63.2/51.0 & 85.3/85.3 & 75.3/70.5 \\
UMDF~\citep{li2024umdf}                  & 85.3/85.2 & 62.9/59.4 & 62.1/62.4 & 85.3/85.3 & 85.3/85.2 & 65.4/62.8 & 85.7/85.6 & 76.0/75.1 \\
CorrKD~\citep{li2024corrkd}              & 85.3/85.1 & 61.9/56.8 & 63.0/62.5 & 85.3/85.2 & 85.6/85.5 & 64.5/62.3 & 85.7/85.6 & 75.9/74.7 \\
MMANet~\citep{wei2023mmanet}             & 84.6/84.6 & 62.9/48.5 & 63.7/63.1 & 84.8/84.8 & 85.3/85.2 & 64.5/60.9 & 85.3/85.2 & 75.9/73.2 \\
CMAD~\citep{zhuang2025cmad}$^{\dagger}$  & 85.7/85.7 & 62.9/58.9 & 63.0/61.6 & 86.7/86.7 & 86.8/86.8 & 63.9/63.0 & 86.8/86.8 & 76.6/75.6 \\
TMDC~\citep{zhuang2026tmdc}$^{\dagger}$  & 85.6/85.6 & 73.5/72.9 & 67.1/67.1 & 85.1/85.2 & 85.5/85.6 & 67.6/67.8 & 84.6/84.7 & 78.4/78.4 \\
MoMKE~\citep{xu2024momke}$^{\dagger}$    & 86.5/86.5 & 71.1/71.1 & 70.2/69.4 & 87.0/87.0 & 87.1/87.0 & 72.7/72.4 & \textbf{87.5}/\textbf{87.4} & 80.3/80.1 \\
\midrule
\multicolumn{9}{@{}l}{\emph{Same frozen models as the three rows above, corrected post hoc under our protocol}}\\
CMAD $+$ GUARD                           & 85.7/85.7 & 67.6/65.1 & 67.6/66.2 & 86.8/86.8 & 86.6/86.6 & 69.7/68.8 & 86.8/86.8 & 78.7/78.0 \\
TMDC $+$ GUARD                           & 85.6/85.6 & \textbf{74.0}/\textbf{73.4} & 72.1/72.0 & 85.5/85.7 & 85.5/85.6 & 72.4/72.4 & 85.5/85.7 & 80.1/80.0 \\
MoMKE $+$ GUARD                          & \textbf{86.7}/\textbf{86.7} & 72.2/72.0 & \textbf{73.0}/\textbf{72.2} & \textbf{87.1}/\textbf{87.1} & \textbf{87.2}/\textbf{87.2} & \textbf{74.9}/\textbf{74.5} & 87.2/87.2 & \textbf{81.2}/\textbf{81.0} \\
\bottomrule
\end{tabular}
\end{table}

We then apply the same policy across the whole suite. Every benchmark mean improves, with joint harm inside the budget. Gain ranges from $19.0$ points where sensor loss exposes large correctable headroom to $1.0$ on a model that largely absorbs the degradation (App.~\ref{app:percond}). Low joint harm need not mean reliable interventions: per benchmark, $17$--$32\%$ of applied corrections still raise the loss beyond the tolerance (App.~\ref{app:conditional-harm}).

\begin{table}[!htbp]
\caption{\textbf{Frozen model and GUARD at $\alpha=0.2$ across six domains.} Metrics follow each benchmark. Harm and apply ranges cover incomplete conditions; CMU-MOSEI uses only language-absent masks ($A$, $V$, $AV$). \emph{Cond.} is the number of degraded conditions. Harm is measured against the frozen output $p_0$ (App.~\ref{app:reference}).}
\label{tab:allbench}
\centering\footnotesize
\setlength{\tabcolsep}{4pt}
\begin{tabular}{@{}llccccc@{}}
\toprule
\textbf{Benchmark (model)} & \textbf{Metric} & \textbf{Cond.} & \textbf{frozen} & \textbf{GUARD} & \textbf{Joint harm} & \textbf{Apply (\%)} \\
\midrule
DrugBAN ($3$ datasets)     & AUROC         & $9$ & $0.718$ & $\mathbf{0.848}$ & $0.049$--$0.165$ & $43$--$74$ \\
OPPORTUNITY (DeepConvLSTM) & wtd.\ F1      & $4$ & $0.563$ & $\mathbf{0.753}$ & $0.148$--$0.186$ & $56$--$71$ \\
CMU-MOSEI (CMAD)           & binary acc.   & $3$ & $0.637$ & $\mathbf{0.688}$ & $0.094$--$0.116$ & $30$--$37$ \\
IEMOCAP (MoMKE)            & $4$-class WA  & $6$ & $0.680$ & $\mathbf{0.690}$ & $0.055$--$0.122$ & $18$--$49$ \\
AVE (AV-att)               & $29$-way acc. & $2$ & $0.464$ & $\mathbf{0.548}$ & $0.177$--$0.178$ & $57$--$65$ \\
NinaPro DB5 (sEMG CNN)     & accuracy      & $10$ & $0.663$ & $\mathbf{0.742}$ & $0.118$--$0.147$ & $75$--$98$ \\
PTB-XL (resnet1d\_wang)    & macro AUC     & $10$ & $0.761$ & $\mathbf{0.852}$ & $0.078$--$0.157$ & $47$--$87$ \\
\bottomrule
\end{tabular}
\end{table}

\subsection{Why not correct every prediction?}
\label{sec:gates}

\begin{figure}[!t]
\centering
\includegraphics[width=\linewidth]{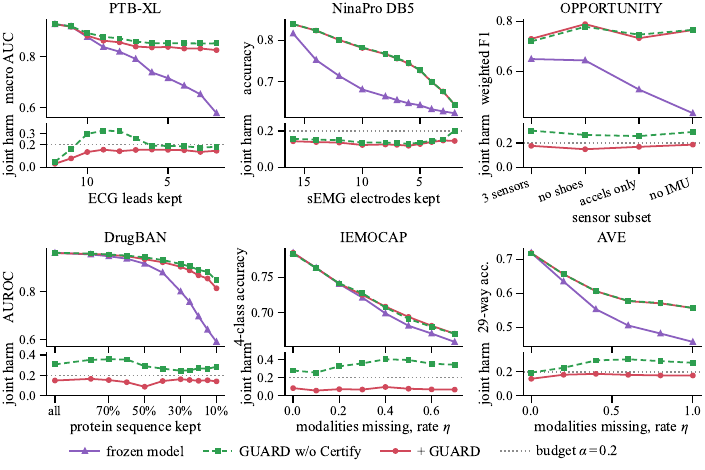}
\caption{\textbf{Utility and harm as the degradation deepens.} Upper axis: the published metric. Lower strip: joint harm against the budget. Dashed \emph{GUARD w/o \textup{\textsc{Certify}}} applies the corrector everywhere and overlaps GUARD where the gate admits everything. Blanket joint harm peaks between $0.202$ on NinaPro and $0.404$ on IEMOCAP, against $0.200$. }%
\label{fig:severity}
\end{figure}

A correction can be useful on average and still be a poor blanket deployment policy. Figure~\ref{fig:severity} isolates this distinction by applying the same corrector either everywhere or only where \textsc{Certify} admits it. The utility curves nearly coincide, but the harm curves separate sharply: the ungated corrector crosses the budget on all six benchmarks, whereas GUARD stays under it on all six.

Table~\ref{tab:gates} fixes the frozen model and corrector on each benchmark and varies only the score, the calibration and the tightening. The corrector supplies the gain: no gated rule exceeds blanket correction by more than $0.001$. Calibration brings joint harm to the budget for every score, even a random one. Realized harm exceeds $\alpha$ only for \emph{GUARD w/o tighten} on CMU-MOSEI ($0.202$) and for the confidence ($0.205$) and random ($0.202$) scores on OPPORTUNITY, all allowed by an expectation bound.

Under the same tightening, the learned score matches or beats confidence and beats a random score on mean gain, with lower mean harm, on all seven benchmarks. The $\ell_1$ distance alone comes close in gain; its lower harm on five benchmarks reflects a lower apply rate. Learn-then-Test records lower harm than conformal risk control on all seven, at a gain cost of at most $0.004$. The two control different probabilities: Learn-then-Test adds a margin for a high-probability statement, and \S\ref{sec:budget} shows what that margin costs when labels are few. Tightening lowers joint harm by $0.018$--$0.096$ for at most $0.006$ gain, except on CMU-MOSEI, where gain falls from $+0.064$ to $+0.051$.

Applied corrections help on balance. Mean loss falls on applied samples on every benchmark, and corrections turn a right decision wrong on $1.3$--$6.4\%$ of samples against $3.0$--$23.8\%$ the other way (Table~\ref{tab:reliability}). At GUARD's own rate, the learned score keeps gain within $0.6$ points of the $\ell_1$ distance and has lower joint and conditional harm on all seven benchmarks, conditional harm by $0.8$--$8.2$ points (Table~\ref{tab:matched}). This matched-rate comparison is diagnostic and carries no guarantee.

\begin{table}[!t]
\caption{\textbf{Scoring, calibration and tightening on one fixed corrector.} Every rule sees the same frozen model, corrector, splits and budget $\alpha=0.2$. \emph{Blanket} corrects everything: an ungated reference, not an upper bound. \emph{Conf.} is the model's confidence~\citep{hendrycks2017msp,geifman2017selective}, \emph{Learned} the score of GUARD, $\ell_1$ the distance between the frozen output and the target, and \emph{Random} a uniform score, all calibrated by conformal risk control. \emph{LTT} applies Learn-then-Test~\citep{angelopoulos2021ltt} to the learned score. The tightening is the same fit-split step in every column of the right block. Cells give accuracy gain\,/\,joint harm rate; a star marks harm above $\alpha$. Details are in App.~\ref{app:setup}.}
\label{tab:gates}
\centering\scriptsize
\setlength{\tabcolsep}{1.5pt}
\resizebox{\linewidth}{!}{%
\begin{tabular}{@{}lccccccccc@{}}
\toprule
 & & \multicolumn{3}{c}{\textbf{Without tightening}} & \multicolumn{5}{c}{\textbf{With tightening}} \\
\cmidrule(lr){3-5}\cmidrule(lr){6-10}
\textbf{Benchmark} & \textbf{Blanket} & Conf. & LTT & Learned & Conf. & $\ell_1$ & Random & LTT & \textbf{GUARD} \\
\midrule
CMU-MOSEI & $+.067$/$.331$$^{\ast}$ & $+.055$/$.198$ & $+.062$/$.177$ & $+.064$/$.202$$^{\ast}$ & $+.043$/$.139$ & $+.043$/$.076$ & $+.029$/$.131$ & $+.048$/$.097$ & $+.051$/$.106$ \\
IEMOCAP & $+.010$/$.315$$^{\ast}$ & $+.010$/$.166$ & $+.010$/$.130$ & $+.009$/$.164$ & $+.008$/$.103$ & $+.009$/$.077$ & $+.004$/$.096$ & $+.008$/$.064$ & $+.008$/$.086$ \\
OPPORTUNITY & $+.183$/$.279$$^{\ast}$ & $+.171$/$.205$$^{\ast}$ & $+.176$/$.171$ & $+.178$/$.187$ & $+.169$/$.200$ & $+.164$/$.156$ & $+.143$/$.202$$^{\ast}$ & $+.176$/$.156$ & $+.178$/$.169$ \\
AVE & $+.084$/$.300$$^{\ast}$ & $+.073$/$.191$ & $+.084$/$.169$ & $+.084$/$.196$ & $+.071$/$.184$ & $+.084$/$.181$ & $+.052$/$.183$ & $+.084$/$.154$ & $+.084$/$.178$ \\
NinaPro DB5 & $+.092$/$.125$ & $+.092$/$.125$ & $+.092$/$.124$ & $+.092$/$.125$ & $+.090$/$.109$ & $+.091$/$.070$ & $+.087$/$.119$ & $+.092$/$.072$ & $+.092$/$.072$ \\
PTB-XL & $+.170$/$.239$$^{\ast}$ & $+.164$/$.193$ & $+.168$/$.169$ & $+.171$/$.194$ & $+.158$/$.175$ & $+.153$/$.156$ & $+.144$/$.180$ & $+.161$/$.138$ & $+.165$/$.150$ \\
DrugBAN & $+.085$/$.275$$^{\ast}$ & $+.074$/$.191$ & $+.084$/$.171$ & $+.085$/$.190$ & $+.072$/$.165$ & $+.078$/$.120$ & $+.048$/$.151$ & $+.084$/$.127$ & $+.084$/$.150$ \\
\bottomrule
\end{tabular}}
\end{table}

\begin{table}[t]
\caption{\textbf{Certified gain under the two retrieval targets.} Targets from \eqref{eq:knn2} share neighbors, with the blend and gate refitted. Accuracy gain is averaged over conditions and seeds; rows are ordered by richer-mask accuracy. Bold marks a gain larger by over two paired standard errors.}
\label{tab:crossmask}
\centering\footnotesize
\setlength{\tabcolsep}{8pt}
\begin{tabular}{lccc}
\toprule
 & \textbf{richer-mask} & \multicolumn{2}{c}{\textbf{gated gain}} \\
\cmidrule(lr){3-4}
\textbf{Dataset, split} & \textbf{accuracy} & \textbf{hard-label} & \textbf{cross-mask} \\
\midrule
Human, random      & $0.931$ & $+0.033$ & $+0.035$ \\
BindingDB, random  & $0.906$ & $+0.146$ & $+0.143$ \\
BioSNAP, random    & $0.829$ & $+0.138$ & $+0.138$ \\
\midrule
OPPORTUNITY        & $0.611$ & $\mathbf{+0.178}$ & $+0.056$ \\
BioSNAP, cluster   & $0.556$ & $\mathbf{+0.112}$ & $+0.028$ \\
BindingDB, cluster & $0.551$ & $\mathbf{+0.234}$ & $+0.028$ \\
\bottomrule
\end{tabular}
\end{table}
\subsection{When can richer-view outputs replace retrieval labels?}
\label{sec:crossmask-main}

The cross-mask result identifies a trade-off: replacing hard labels removes retrieval-label noise but
inherits error from the richer-view predictor. Table~\ref{tab:crossmask} studies its practical effect by holding the neighbors fixed and changing the target before fitting the blend and gate. When the richer predictor is
strong, cross-mask transfer matches hard-label retrieval to within $0.003$ without
retrieval labels, also from a pool the model never trained on (App.~\ref{app:pool}). When it is weak ($0.551$--$0.611$ accuracy), most of the gain is lost. Proposition~\ref{prop:transfer} is conditional, so this is not a universal accuracy threshold; conditional correctness is not identifiable from unlabeled data (App.~\ref{app:screen}).

\subsection{How many labels does it need, and what if deployment changes?}
\label{sec:budget}

Table~\ref{tab:budget} fixes the test split and varies the total number of labeled deployment samples. On CMU-MOSEI, $480$ labels bring cross-mask retrieval to $+0.036$, more than hard retrieval reaches with all $3483$ ($+0.029$). On OPPORTUNITY, whose richer model is weak, cross-mask stalls near $+0.05$ and hard retrieval pulls away from $960$ labels. On BioSNAP cluster, cross-mask leads up to $480$ labels, then hard targets lead and the probe ends above retrieval. No corrector wins everywhere; the label-free target pays when labels are few. Small hard-label budgets can give a slightly negative gain on CMU-MOSEI: the budget bounds harm, not accuracy.

When the pool already carries training labels, $40$ deployment labels, half for fitting and half for calibration, keep $78\%$ of PTB-XL's gain and $64\%$ of AVE's (App.~\ref{app:budget}). The calibrators differ most at few labels: across the $102$ budget settings of these studies, GUARD has the higher gain in $85$ and the lower in $9$, at higher harm inside the budget, by a mean $1.7$ points with at most $25$ calibration labels and $0.25$ with more than $350$.

\begin{table}[!htbp]
\caption{\textbf{Accuracy gain of GUARD under a total budget of labeled deployment samples.} Hard targets split the budget equally across $D_{\mathrm{source}}$, $D_{\mathrm{fit}}$ and $D_{\mathrm{conf}}$. Cross-mask targets keep the whole unlabeled pool and split it across the last two. \emph{Full} uses every role in full. Mean joint harm stays below $\alpha$ in every cell.}
\label{tab:budget}
\centering\scriptsize
\setlength{\tabcolsep}{2.4pt}
\begin{tabular}{@{}lcccccccccccc@{}}
\toprule
 & \multicolumn{4}{c}{\textbf{CMU-MOSEI}} & \multicolumn{4}{c}{\textbf{OPPORTUNITY}} & \multicolumn{4}{c}{\textbf{BioSNAP cluster}} \\
\cmidrule(lr){2-5}\cmidrule(lr){6-9}\cmidrule(lr){10-13}
 & \multicolumn{2}{c}{Retrieval} & \multicolumn{2}{c}{Probe} & \multicolumn{2}{c}{Retrieval} & \multicolumn{2}{c}{Probe} & \multicolumn{2}{c}{Retrieval} & \multicolumn{2}{c}{Probe} \\
\textbf{Labels} & hard & cross & hard & cross & hard & cross & hard & cross & hard & cross & hard & cross \\
\midrule
$60$ & $-.012$ & $+.017$ & $-.003$ & $+.020$ & $+.028$ & $+.021$ & $+.027$ & $+.021$ & $+.018$ & $+.038$ & $+.020$ & $-.006$ \\
$120$ & $-.007$ & $+.027$ & $-.007$ & $+.006$ & $+.029$ & $+.026$ & $+.034$ & $+.030$ & $+.029$ & $+.044$ & $+.014$ & $+.007$ \\
$240$ & $-.003$ & $+.025$ & $-.003$ & $+.012$ & $+.035$ & $+.032$ & $+.052$ & $+.042$ & $+.015$ & $+.051$ & $+.008$ & $+.006$ \\
$480$ & $+.002$ & $+.036$ & $+.001$ & $+.022$ & $+.061$ & $+.058$ & $+.058$ & $+.048$ & $+.043$ & $+.050$ & $+.032$ & $+.014$ \\
$960$ & $-.001$ & $+.045$ & $+.002$ & $+.023$ & $+.128$ & $+.048$ & $+.101$ & $+.050$ & $+.058$ & $+.051$ & $+.067$ & $+.021$ \\
full & $+.029$ & $+.050$ & $+.015$ & $+.026$ & $+.178$ & $+.056$ & $+.135$ & $+.050$ & $+.101$ & $+.057$ & $+.131$ & $+.026$ \\
\bottomrule
\end{tabular}
\end{table}

A deployment can also change. Under mixed availability, one threshold controls harm on the mixture but not on each pattern: without tightening, the worst pattern reaches $0.29$ on IEMOCAP. One threshold per mask (Cor.~\ref{cor:mask}) lowers the worst pattern to at most $0.19$ there and from $0.24$ to $0.207\pm0.008$ on CMU-MOSEI, at the same gain. Calibration data from another condition fall outside the guarantee, though mean harm stayed inside the budget here when the corrector was fitted at the deployed condition. Reusing the whole adapter fails: harm reaches $0.23$ on CMU-MOSEI, $0.22$ on IEMOCAP and $0.52$ on NinaPro, and gain turns negative (App.~\ref{app:shift}). The mask is observed, so each condition can keep its own adapter; drift within a mask falls outside the guarantee.

\section{Discussion and Conclusion}

Missing modalities leave a frozen model with a Bayes gap that the observed inputs can reduce and ambiguity that they cannot; richer-view outputs can replace retrieval labels when the richer model is conditionally correct. Because a helpful correction can still hurt individual predictions, GUARD separates building a candidate from deciding when to apply it, keeping mean joint harm within budget on every benchmark. Its guarantee is marginal, assumes exchangeability and needs labeled deployment data, and an adapter fitted at one condition does not carry over to another. Except for the cross-mask target, the analysis covers any frozen model whose inputs change. Conditional risk control, shift-aware calibration and fewer labels could make certified correction a general maintenance layer for deployed models.

\bibliography{guard_preprint}
\bibliographystyle{iclr2027_conference}

\clearpage
\appendix
\addtocontents{toc}{\protect\setcounter{tocdepth}{2}}

\setlength{\parskip}{0pt}
\renewcommand{\contentsname}{Appendix: Table of Contents}
\setcounter{tocdepth}{2}
\tableofcontents
\clearpage

\makeatletter
\renewcommand\@fb@secFB{}
\setlength{\@fpsep}{14pt}
\setlength{\@fpbot}{0pt plus 1fil}
\makeatother

\section{Notation}
\label{app:guide}

Table~\ref{tab:notation} collects the notation. The rest of this appendix is organized as follows. App.~\ref{app:setup} details the experimental protocol: frozen models, data roles, search grids and baselines. App.~\ref{app:percond} gives the per-condition results behind the summary tables of \S\ref{sec:experiments}. App.~\ref{app:analyses} collects further analyses of the gate, the corrector, the action score, the data requirements and the budget. App.~\ref{app:synthetic} describes the synthetic family behind Figure~\ref{fig:synthetic}. App.~\ref{sec:limitations} states where each result applies, and App.~\ref{app:proof-reduction} onward contains the proofs.

\begin{table}[!htbp]
\caption{Notation used throughout the paper and this appendix.}
\label{tab:notation}
\centering\footnotesize
\begin{tabular}{@{}ll@{}}
\toprule
\textbf{Symbol} & \textbf{Meaning} \\
\midrule
$Y$ & the label \\
$S\subseteq T$ & modality masks; $S$ observes a subset of what $T$ observes \\
$X=(x_S,S)$ & everything about a sample the corrector may read at deployment \\
$z(X)$ & the frozen model's scores \\
$\ell(y,z)$, $\bar\ell(y,p)$ & score-space loss and prediction loss, respectively \\
$m(z)$ & the prediction in mean-parameter form; $\softmax(z)$ for cross-entropy \\
$t(y)$ & the label statistic; $e_y$ for cross-entropy, $y$ for squared error \\
$\pi(X)$ & $\E[t(Y)\mid X]$, the best prediction the observed inputs allow \\
$d(X)$ & $m(z(X))-\pi(X)$, the \emph{Bayes gap}: the part of the error a corrector can reach \\
$\pi_S,\,d_S,\,m_S$ & the same quantities under mask $S$ \\
$b$ & $\E[d_T\mid X_S]$, the richer mask's gap inherited by the poorer one \\
$\widehat\pi$ & the frozen plug-in estimate of $\pi$, built from $D_{\mathrm{source}}$ \\
$k$, $\mathcal N_k(X)$ & neighborhood size and the retrieved neighbors \\
$p_0$, $p_{\base}$ & the frozen output $m(z)$, and the harm reference/fallback, equal to $p_0$ throughout \\
$p_{\mathrm{temp}}$, $p_{\corr}$ & temperature-scaled frozen output, and the candidate $(1-\widehat\beta)p_{\mathrm{temp}}+\widehat\beta\widehat\pi$ \\
$\vartheta$ & temperature; $\vartheta=1$ leaves probabilities unchanged \\
$\beta$, $\widehat\beta$ & blend weight in $[0,1]$, and its fitted value \\
$r_\theta(X)$ & learned action score, fitted on $D_{\mathrm{fit}}$ \\
$A_\lambda,\ L_\lambda,\ \widehat\lambda$ & threshold policy, its harmful-intervention loss, and the threshold calibrated by CRC \\
$\Delta_Y$, $\alpha$, $\delta$ & loss change $\bar\ell(Y,p_{\corr})-\bar\ell(Y,p_{\base})$; harm budget and tolerance, $0.2$ and $0.05$ \\
$D_{\mathrm{source}}$ & builds the target estimator \\
$D_{\mathrm{fit}}$ & selects $k$, target, space, weighting, temperature and $\widehat\beta$ \\
$D_{\mathrm{conf}}$, $D_{\mathrm{test}}$ & calibrates the gate; read once after every choice is fixed \\
\bottomrule
\end{tabular}
\end{table}

\section{Experimental Setup}
\label{app:setup}

Each condition assigns disjoint roles to $D_{\mathrm{source}}$, $D_{\mathrm{fit}}$, $D_{\mathrm{conf}}$ and $D_{\mathrm{test}}$, and every decision rule is frozen before test evaluation. Unless stated otherwise, $\alpha=0.2$, $\delta=0.05$, and $\widehat\beta$ is chosen on $D_{\mathrm{fit}}$ by the loss-minimizing rule of \S\ref{sec:recal}. The neighborhood size $k$, the retrieval target, the retrieval space, the neighbor weighting and the temperature are selected on $D_{\mathrm{fit}}$ by the metric the benchmark reports. The rule comparison of Table~\ref{tab:gates} is the exception and selects on accuracy throughout. The test split is read once, after selection.

\paragraph{Frozen models.}
Table~\ref{tab:setup} lists the nine frozen model--dataset pairs. Each model is anchored to its published number before any correction is attached. Every model we correct is published except NinaPro's sEMG CNN, which is our own and follows the dataset paper's protocol~\citep{pizzolato2017ninapro}. Only AVE ships usable weights~\citep{tian2018ave}; we retrain the others from the authors' released code~\citep{bai2023drugban,ordonez2016deepconvlstm,strodthoff2021ptbxl,xu2024momke,zhuang2025cmad}. Two anchors fall short: DrugBAN on the BioSNAP \emph{cluster} split, and MoMKE under sparse availability patterns. Results on these two models should be read against our own frozen rows rather than the published tables.

\begin{table}[!htbp]
\caption{\textbf{Benchmarks, frozen models, and degradations.} \emph{Anchor} compares each frozen model against its published number before any correction is attached; where a model is published across many availability patterns we give the worst gap over all of them, and Human has no directly comparable published entry. $^{\ast}$\,The only released weights we could use. $^{\dagger}$\,Our own model, so its anchor is the dataset paper's protocol rather than a reproduction.}
\label{tab:setup}
\centering\footnotesize
\setlength{\tabcolsep}{4pt}
\begin{tabular}{@{}llll@{}}
\toprule
\textbf{Dataset} & \textbf{Frozen model} & \textbf{Anchor: published $\to$ ours} & \textbf{Degradations} \\
\midrule
BioSNAP     & DrugBAN            & AUROC $.903\!\to\!.904$        & protein $25\%$, $50\%$; scaffold \\
BindingDB   & DrugBAN            & AUROC $.960\!\to\!.963$        & protein $25\%$, $50\%$; scaffold \\
Human       & DrugBAN            & ---                            & protein $25\%$, $50\%$ \\
OPPORTUNITY & DeepConvLSTM       & wtd.\ F1 $.895\!\to\!.879$     & four sensor subsets \\
CMU-MOSEI   & CMAD               & acc.\ within $.015$            & audio, visual, or both \\
IEMOCAP     & MoMKE              & WA $.801\!\to\!.778$ full; short sparse & audio only \\
AVE         & AV-att$^{\ast}$    & acc.\ $.7270\!\to\!.7274$      & audio only; visual only \\
NinaPro DB5 & sEMG CNN$^{\dagger}$ & acc.\ $.690\!\to\!.814$      & $16$ down to $2$ of $16$ electrodes \\
PTB-XL      & resnet1d\_wang     & mAUC $.930\!\to\!.927$         & $11$ down to $2$ leads, precordial first \\
\bottomrule
\end{tabular}
\end{table}

\paragraph{Data roles and search grids.}
Table~\ref{tab:splits} gives the size of each data role and the grids searched on $D_{\mathrm{fit}}$. IEMOCAP is split by video, OPPORTUNITY by contiguous time block, and the other benchmarks by row within the evaluated population. Exchangeability is therefore claimed for held-out rows or blocks, not for an unseen subject.

\begin{table}[!htbp]
\caption{Sample counts per data role, and the grids searched on $D_{\mathrm{fit}}$. Roles are
disjoint at the row level. Counts are per condition; IEMOCAP deployment roles are split by video and OPPORTUNITY roles by contiguous blocks, while the other deployment subdivisions are by row within the evaluated population. OPPORTUNITY counts are means over seeds; IEMOCAP counts are means over its
five folds, whose evaluation sets range from $1023$ to $1241$ samples; NinaPro counts are per subject.}
\label{tab:splits}
\centering\footnotesize
\setlength{\tabcolsep}{5pt}
\begin{tabular}{@{}lrrrrl@{}}
\toprule
\textbf{Benchmark} & $|D_{\mathrm{source}}|$ & $|D_{\mathrm{fit}}|$ & $|D_{\mathrm{conf}}|$ & $|D_{\mathrm{test}}|$ & \textbf{Construction} \\
\midrule
CMU-MOSEI            & $1161$  & $1161$ & $1161$ & $1160$ & four disjoint quarters \\
IEMOCAP              & $4425$  & $436$  & $344$  & $325$  & pool from other folds; fold split by video \\
AVE                  & $33280$ & $1340$ & $1340$ & $1340$ & released pool; deployment split in thirds \\
NinaPro              & $14259$ & $2237$ & $2237$ & $2238$ & pool from other repetitions; rest in thirds \\
OPPORTUNITY          & $2772$  & $2735$ & $2574$ & $2500$ & random $200$-window blocks \\
PTB-XL               & $17084$ & $719$  & $719$  & $720$  & official patient folds \\
DrugBAN (BindingDB)  & $12000$ & $2460$ & $2460$ & $9840$ & released split \\
\midrule
\multicolumn{6}{@{}l}{\emph{Grids searched on $D_{\mathrm{fit}}$, scored by the metric the benchmark reports}} \\
Neighborhood $k$    & \multicolumn{5}{l}{$\{3,5,8,12,20,35,50\}$} \\
Retrieval target     & \multicolumn{5}{l}{hard label, cross-mask} \\
Retrieval space      & \multicolumn{5}{l}{standardized, cosine} \\
Neighbor weighting  & \multicolumn{5}{l}{uniform, distance kernel} \\
Temperature          & \multicolumn{5}{l}{$1.0$, $2.0$} \\
Blend weight $\beta$ & \multicolumn{5}{l}{$41$ equally spaced values on $[0,1]$, by loss on $D_{\mathrm{fit}}$} \\
\bottomrule
\end{tabular}
\end{table}

\paragraph{Prediction loss and temperature reference.}
\label{app:reference}
Write $p_0=m(z)$ for the original frozen output. Models that save logits, as MoMKE does, are converted by their link function first. For categorical predictions, $\operatorname{Temp}_{\vartheta}(p)_c=p_c^{1/\vartheta}/\sum_jp_j^{1/\vartheta}$; for multi-label outputs it applies $\operatorname{sigmoid}(\operatorname{logit}(p_c)/\vartheta)$ coordinatewise. The selected temperature may also be applied to the richer-view retrieval values. All such choices are fixed before calibration.

Every evaluation in the paper uses $p_{\base}=p_0$: the temperature belongs to the candidate, while the action features, the harm event and the prediction a rejected query keeps all read the raw frozen output. Positive temperature scaling preserves class decisions and rankings, so it leaves every frozen accuracy, F1 and AUROC entry unchanged. For the same reason, temperature scaling alone gains nothing under any metric reported here.

\paragraph{Degradation axes.}
PTB-XL's lead ladder removes precordial before limb leads and extends the reduced-lead configurations of the PhysioNet/CinC challenge~\citep{reyna2021cinc}. NinaPro's electrode ladder follows a literature that asks how few sEMG channels a gesture classifier can be built on, which reports usable accuracy well below the full array~\citep{naik2016semg,pelaez2022channels,simon2024channels}; we sweep the range rather than adopt one recommended count. Drug--target interaction has no established missing-modality protocol, so its degradations were fixed in advance. The remaining axes are the benchmarks' own availability patterns, in the spirit of tolerance evaluations such as \citet{zheng2026mutebench}. The panels of Figure~\ref{fig:severity} sweep, in turn, the lead ladder, the electrode budget, OPPORTUNITY's four sensor subsets, the fraction of target protein retained on BindingDB (scored by AUROC), and the deployment missing rate on IEMOCAP and AVE.

\paragraph{Baselines.} We compare against two families. \emph{Without a guarantee}: the
frozen model itself; blanket correction; the confidence scores standard in selective
prediction, namely maximum softmax probability~\citep{hendrycks2017msp}, its margin, and
negative entropy~\citep{geifman2017selective}; agreement between the model and the
retrieval target; a gate fitted on $D_{\mathrm{fit}}$ to predict whether correction helps,
which transposes the per-sample routing principle of \citet{gao2026sieve} to a frozen
model; and published missing-modality methods quoted from their own tables. \emph{With a guarantee}: two rules carry a finite-sample harm
guarantee, possible because harmful application is observable on labeled calibration data. \emph{Conf.\ score} applies conformal risk control to the model's own confidence, and \emph{LTT calib.} calibrates a learned logistic score by Learn-then-Test~\citep{angelopoulos2021ltt}, with Hoeffding--Bentkus $p$-values tested in a fixed order chosen on $D_{\mathrm{fit}}$, stopping at the first non-rejection, at significance level $\zeta=0.05$, which is a distinct quantity from the tolerated loss increase $\delta$. Table~\ref{tab:gates}
therefore compares guarantees as well as decision rules; what it cannot compare is the published
missing-modality methods, which are quoted from their own tables and carry no such statement.

\paragraph{Comparators of Table~\ref{tab:gates}.} The two calibrations control different things: conformal risk control bounds the risk in expectation over the calibration draw, Learn-then-Test bounds the risk of the threshold it selects with probability $1-\zeta$ under its i.i.d.\ sampling assumption ($\zeta=0.05$), and neither bounds the harm realized on a given split. Every column of the table reuses one corrector, and the learned score is the deployed one on every benchmark. Because the table scores every benchmark in accuracy, its numbers differ from the published metrics of Table~\ref{tab:allbench}: OPPORTUNITY's weighted-F1 gain is $+.190$ against the $+.178$ in accuracy shown there, DrugBAN is reported elsewhere by AUROC and PTB-XL by macro AUC.

\section{Per-Condition Results}
\label{app:percond}

Tables~\ref{tab:main}--\ref{tab:fourdomains} give every condition behind Table~\ref{tab:allbench}. Where available, an \emph{intact} row feeds the undegraded input back in. It shows how much headroom a frozen model retains when nothing is missing, which Theorem~\ref{thm:reduction} does not require to be zero.

\begin{table}[!htbp]
\caption{\textbf{DrugBAN under degraded inputs.} AUROC and accuracy for the frozen model and for GUARD, on the two in-domain \emph{random} splits and the cross-domain \emph{cluster} protocol defined by the model's authors. Mean over three seeds, except the two BindingDB \emph{cluster} rows, which are single-seed. $^{\dagger}$\,Protein truncated to $25\%$; $^{\ddagger}$\,Murcko scaffold with the protein truncated to $50\%$. The \emph{random} rows retrieve from the source population defined by the released split; the \emph{cluster} rows retrieve from the deployment domain (App.~\ref{app:pool}).}
\label{tab:main}
\centering
\footnotesize
\setlength{\tabcolsep}{6pt}
\begin{tabular}{llcccccc}
\toprule
 & & \multicolumn{2}{c}{\textbf{AUROC}} & \multicolumn{2}{c}{\textbf{Accuracy}} & & \\
\cmidrule(lr){3-4}\cmidrule(lr){5-6}
\textbf{Dataset} & \textbf{Split} & frozen & GUARD & frozen & GUARD & \textbf{Joint harm} & \textbf{Apply (\%)} \\
\midrule
BioSNAP   & random$^{\dagger}$    & $0.680$ & $\mathbf{0.808}$ & $0.571$ & $\mathbf{0.746}$ & $0.165$ & $54$ \\
          & random$^{\ddagger}$   & $0.750$ & $\mathbf{0.836}$ & $0.675$ & $\mathbf{0.780}$ & $0.159$ & $49$ \\
          & cluster$^{\dagger}$   & $0.560$ & $\mathbf{0.635}$ & $0.505$ & $\mathbf{0.602}$ & $0.156$ & $43$ \\
\midrule
BindingDB & random$^{\dagger}$    & $0.758$ & $\mathbf{0.887}$ & $0.678$ & $\mathbf{0.835}$ & $0.142$ & $71$ \\
          & random$^{\ddagger}$   & $0.800$ & $\mathbf{0.908}$ & $0.722$ & $\mathbf{0.853}$ & $0.163$ & $67$ \\
          & cluster$^{\dagger}$   & $0.503$ & $\mathbf{0.790}$ & $0.529$ & $\mathbf{0.731}$ & $0.137$ & $70$ \\
          & cluster$^{\ddagger}$  & $0.522$ & $\mathbf{0.850}$ & $0.528$ & $\mathbf{0.794}$ & $0.111$ & $74$ \\
\midrule
Human     & random$^{\dagger}$    & $0.948$ & $\mathbf{0.961}$ & $0.869$ & $\mathbf{0.902}$ & $0.049$ & $59$ \\
          & random$^{\ddagger}$   & $0.944$ & $\mathbf{0.960}$ & $0.861$ & $\mathbf{0.896}$ & $0.108$ & $62$ \\
\bottomrule
\end{tabular}
\end{table}

\begin{table}[!htbp]
\caption{\looseness=-1 \textbf{OPPORTUNITY under sensor loss.} Weighted F1, the metric DeepConvLSTM is published with, and accuracy, for the frozen model and for GUARD over four sensor configurations, mean over three seeds.}
\label{tab:gateablation}
\centering\footnotesize
\setlength{\tabcolsep}{7pt}
\begin{tabular}{lccccccc}
\toprule
 & \multicolumn{2}{c}{\textbf{Weighted F1}} & \multicolumn{2}{c}{\textbf{Accuracy}} & & \\
\cmidrule(lr){2-3}\cmidrule(lr){4-5}
\textbf{Condition} & frozen & GUARD & frozen & GUARD & \textbf{Joint harm} & \textbf{Apply (\%)} \\
\midrule
low-cost accels only  & $0.527$ & $\mathbf{0.732}$ & $0.538$ & $\mathbf{0.741}$ & $0.168$ & $64$ \\
no IMU family         & $0.433$ & $\mathbf{0.764}$ & $0.502$ & $\mathbf{0.783}$ & $0.186$ & $69$ \\
no shoe sensors       & $0.643$ & $\mathbf{0.788}$ & $0.646$ & $\mathbf{0.796}$ & $0.148$ & $56$ \\
three sensors only    & $0.649$ & $\mathbf{0.729}$ & $0.660$ & $\mathbf{0.735}$ & $0.176$ & $71$ \\
\bottomrule
\end{tabular}
\end{table}

\begin{table}[!htbp]
\caption{\textbf{Certified correction on five further benchmarks.} The frozen model against GUARD at $\alpha=0.2$, each block scored with the metric of its dataset, named beneath the dataset. CMU-MOSEI binary accuracy is computed on non-neutral samples ($y\neq0$), the standard protocol for that dataset, and averages ten deployment splits, not the five of Table~\ref{tab:mosei-main}.}
\label{tab:fourdomains}
\centering\footnotesize
\setlength{\tabcolsep}{6pt}
\begin{tabular}{llcccc}
\toprule
\textbf{Dataset (model)} & \textbf{Available} & frozen & GUARD & \textbf{Joint harm} & \textbf{Apply (\%)} \\
\midrule
CMU-MOSEI (CMAD)      & audio $+$ visual    & $0.645$ & $\mathbf{0.703}$ & $0.116$ & $37$ \\
\emph{binary accuracy} & audio only          & $0.631$ & $\mathbf{0.679}$ & $0.094$ & $30$ \\
                      & visual only         & $0.637$ & $\mathbf{0.683}$ & $0.107$ & $32$ \\
                      & \emph{intact}       & $0.863$ & $\mathbf{0.864}$ & $0.027$ & $25$ \\
\midrule
IEMOCAP (MoMKE)       & audio only & $0.617$ & $\mathbf{0.643}$ & $0.122$ & $49$ \\
\emph{$4$-class WA}   & text only & $0.745$ & $\mathbf{0.758}$ & $0.102$ & $40$ \\
                      & visual only & $\mathbf{0.547}$ & $0.546$ & $0.055$ & $18$ \\
                      & audio $+$ text & $0.793$ & $\mathbf{0.797}$ & $0.057$ & $21$ \\
                      & audio $+$ visual & $0.629$ & $\mathbf{0.641}$ & $0.092$ & $25$ \\
                      & text $+$ visual & $0.750$ & $\mathbf{0.754}$ & $0.090$ & $27$ \\
                      & \emph{intact} & $0.785$ & $0.785$ & $0.085$ & $29$ \\
\midrule
AVE (AV-att)          & audio only          & $0.465$ & $\mathbf{0.600}$ & $0.177$ & $65$ \\
\emph{$29$-way acc.}  & visual only         & $0.462$ & $\mathbf{0.496}$ & $0.178$ & $57$ \\
                      & \emph{intact}       & $\mathbf{0.719}$ & $0.718$ & $0.169$ & $85$ \\
\midrule
NinaPro DB5 (sEMG CNN) & $14$ electrodes     & $0.752$ & $\mathbf{0.824}$ & $0.139$ & $97$ \\
\emph{accuracy}       & $12$ electrodes     & $0.713$ & $\mathbf{0.801}$ & $0.136$ & $97$ \\
                      & $10$ electrodes     & $0.680$ & $\mathbf{0.781}$ & $0.122$ & $97$ \\
                      & $8$ electrodes      & $0.664$ & $\mathbf{0.767}$ & $0.125$ & $97$ \\
                      & $7$ electrodes      & $0.654$ & $\mathbf{0.756}$ & $0.123$ & $97$ \\
                      & $6$ electrodes      & $0.647$ & $\mathbf{0.744}$ & $0.118$ & $97$ \\
                      & $5$ electrodes      & $0.642$ & $\mathbf{0.727}$ & $0.128$ & $98$ \\
                      & $4$ electrodes      & $0.632$ & $\mathbf{0.699}$ & $0.139$ & $98$ \\
                      & $3$ electrodes      & $0.627$ & $\mathbf{0.675}$ & $0.147$ & $98$ \\
                      & $2$ electrodes      & $0.623$ & $\mathbf{0.643}$ & $0.145$ & $75$ \\
                      & \emph{intact}       & $0.817$ & $\mathbf{0.839}$ & $0.143$ & $97$ \\
\midrule
PTB-XL (resnet1d\_wang) & $11$ leads          & $\mathbf{0.919}$ & $0.918$ & $0.078$ & $47$ \\
\emph{macro AUC}      & $10$ leads          & $0.877$ & $\mathbf{0.882}$ & $0.135$ & $62$ \\
                      & $9$ leads           & $0.838$ & $\mathbf{0.863}$ & $0.156$ & $68$ \\
                      & $8$ leads           & $0.819$ & $\mathbf{0.856}$ & $0.143$ & $63$ \\
                      & $7$ leads           & $0.791$ & $\mathbf{0.840}$ & $0.154$ & $70$ \\
                      & $6$ leads           & $0.738$ & $\mathbf{0.836}$ & $0.157$ & $87$ \\
                      & $5$ leads           & $0.715$ & $\mathbf{0.838}$ & $0.154$ & $87$ \\
                      & $4$ leads           & $0.685$ & $\mathbf{0.832}$ & $0.151$ & $86$ \\
                      & $3$ leads           & $0.652$ & $\mathbf{0.832}$ & $0.136$ & $86$ \\
                      & $2$ leads           & $0.578$ & $\mathbf{0.826}$ & $0.146$ & $84$ \\
                      & \emph{intact}       & $0.927$ & $\mathbf{0.928}$ & $0.031$ & $76$ \\
\bottomrule
\end{tabular}
\end{table}

\paragraph{Fitted blend weight on CMU-MOSEI.} Over the three frozen models and seven modality subsets, $\widehat\beta$ averages $0.70$ where language is missing against $0.12$ where it is present, so the correction is scaled down rather than switched off: it is selected to zero on only two of the thirty-three language-keeping cells. The average gain ranges from $2.1$ accuracy points on CMAD to $0.9$ on MoMKE.

\section{Additional Analyses}
\label{app:analyses}

\subsection{Removing the gate}

Table~\ref{tab:blanket} compares GUARD with blanket correction, the same corrector applied to every sample, over the conditions of Table~\ref{tab:gates}. Averaged over conditions, blanket joint harm is $1.6$ to $3.6$ times that of GUARD, and its range exceeds the budget on six of the seven benchmarks. The two policies are closest at mild degradation, where the frozen model has little to correct.

\begin{table}[!htbp]
\caption{\textbf{Effect of removing the gate.} Blanket correction applies the frozen corrector of GUARD to every sample. Joint harm is the range over each benchmark's conditions, against the budget $\alpha=0.2$. $\Delta$ is the accuracy gain of blanket correction, with per-label accuracy on PTB-XL. Conditions follow Table~\ref{tab:gates}, so ranges can differ from those of Table~\ref{tab:allbench}.}
\label{tab:blanket}
\centering\footnotesize
\setlength{\tabcolsep}{7pt}
\begin{tabular}{lccccc}
\toprule
 & & \multicolumn{2}{c}{\textbf{Blanket correction}} & \textbf{GUARD} \\
\cmidrule(lr){3-4}\cmidrule(lr){5-5}
\textbf{Benchmark} & \textbf{Conditions} & \textbf{$\Delta$} & \textbf{joint harm} & \textbf{joint harm} \\
\midrule
CMU-MOSEI    & $3$ & $+0.067$ & $0.302$--$0.351$ & $\mathbf{0.094}$--$\mathbf{0.116}$ \\
IEMOCAP      & $7$ & $+0.010$ & $0.206$--$0.368$ & $\mathbf{0.055}$--$\mathbf{0.122}$ \\
OPPORTUNITY  & $4$ & $+0.183$ & $0.256$--$0.302$ & $\mathbf{0.148}$--$\mathbf{0.186}$ \\
AVE          & $2$ & $+0.084$ & $0.291$--$0.310$ & $\mathbf{0.177}$--$\mathbf{0.178}$ \\
NinaPro DB5  & $4$ & $+0.092$ & $0.109$--$0.153$ & $\mathbf{0.066}$--$\mathbf{0.084}$ \\
PTB-XL       & $6$ & $+0.170$ & $0.175$--$0.329$ & $\mathbf{0.126}$--$\mathbf{0.164}$ \\
DrugBAN      & $3$ & $+0.085$ & $0.243$--$0.290$ & $\mathbf{0.135}$--$\mathbf{0.187}$ \\
\bottomrule
\end{tabular}
\end{table}

\subsection{Harm among applied interventions}
\label{app:conditional-harm}

A small joint harm rate need not mean that most applied corrections help. For a fixed policy with positive apply probability, conditional harm equals joint harm divided by apply probability. If the apply probability is at most $\alpha$, the joint constraint alone permits every applied correction to be harmful. Table~\ref{tab:reliability} reports this rate, together with what applied corrections do to the decision.

\begin{table}[!htbp]
\caption{\textbf{What applied corrections do.} GUARD at $\alpha=0.2$, over the cells of Table~\ref{tab:gates}. \emph{Cond.\ harm} is joint harm over the rate of applications that change the output; \emph{worst} is its largest condition. $\Delta$ is the mean loss change on applied samples. Flip$^-$ and flip$^+$ are the shares of samples on which an applied correction turns a correct decision wrong or a wrong one correct.}
\label{tab:reliability}
\centering\footnotesize
\setlength{\tabcolsep}{8pt}
\begin{tabular}{@{}lccccc@{}}
\toprule
 & \multicolumn{2}{c}{Cond.\ harm (\%)} & & & \\
\cmidrule(lr){2-3}
Benchmark & mean & worst & $\Delta$ & Flip$^-$ & Flip$^+$ \\
\midrule
CMU-MOSEI & $32.2$ & $33.9$ & $-0.06$ & $0.040$ & $0.090$ \\
IEMOCAP & $29.1$ & $36.9$ & $-0.08$ & $0.022$ & $0.030$ \\
OPPORTUNITY & $26.1$ & $27.0$ & $-2.81$ & $0.060$ & $0.238$ \\
AVE & $29.2$ & $31.4$ & $-3.24$ & $0.031$ & $0.115$ \\
NinaPro DB5 & $17.3$ & $18.6$ & $-4.49$ & $0.013$ & $0.104$ \\
PTB-XL & $19.4$ & $24.2$ & $-0.58$ & $0.064$ & $0.229$ \\
DrugBAN & $23.9$ & $44.8$ & $-0.55$ & $0.054$ & $0.138$ \\
\bottomrule
\end{tabular}
\end{table}

Per benchmark, $17$--$32\%$ of applied corrections raise the loss beyond the tolerance. This harmful share is not a net loss. Mean loss falls on applied samples everywhere, and corrections turn wrong decisions right more often than the reverse, by $1.4$ times on IEMOCAP and up to $8$ times on NinaPro. The learned score ranks where correction helps; it is not trained to avoid the harm event. Label ambiguity also limits how far any gate can go: for binary labels, among corrections that exceed the tolerance, the harm rate is at least the Bayes error of those samples (Prop.~\ref{prop:floor}). An application that needs reliable individual interventions would need a separately calibrated conditional-risk policy; lowering the joint budget alone does not provide one.

Table~\ref{tab:matched} compares ranking rules at GUARD's own intervention rate. Every rule applies the same corrector to its top-ranked test samples, so the rate is matched by construction and no rule carries a guarantee. Against the $\ell_1$ distance, the learned score keeps gain within $0.6$ points and has lower joint and conditional harm on all seven benchmarks.

\begin{table}[!htbp]
\caption{\textbf{Rules at GUARD's own intervention rate.} One fixed corrector per benchmark. In each run every rule applies it to its top-ranked test samples, up to the share GUARD applies there, so the rate is matched by construction and no rule here carries a guarantee. Cells give gain/joint harm/conditional harm (\%). $\ell_1$ ranks by the distance between the frozen output and the retrieval target. Agreement is a binary flag, so its ties are arbitrary; it is constant on AVE and NinaPro.}
\label{tab:matched}
\centering\scriptsize
\setlength{\tabcolsep}{2.5pt}
\begin{tabular}{@{}lcccccc@{}}
\toprule
Benchmark & Apply (\%) & GUARD & $\ell_1$ & Confidence & Agreement & Random \\
\midrule
CMU-MOSEI & $32.9$ & $+.051$/$.106$/$32.2$ & $+.056$/$.127$/$38.7$ & $+.034$/$.105$/$31.8$ & $+.007$/$.101$/$30.6$ & $+.025$/$.108$/$32.9$ \\
IEMOCAP & $29.7$ & $+.008$/$.086$/$29.1$ & $+.009$/$.105$/$35.6$ & $+.009$/$.095$/$32.0$ & $+.001$/$.087$/$29.4$ & $+.004$/$.094$/$31.6$ \\
OPPORTUNITY & $65.0$ & $+.178$/$.169$/$26.1$ & $+.171$/$.212$/$32.5$ & $+.149$/$.185$/$28.4$ & $+.087$/$.159$/$24.5$ & $+.122$/$.180$/$27.7$ \\
AVE & $60.8$ & $+.084$/$.178$/$29.2$ & $+.084$/$.185$/$30.3$ & $+.072$/$.193$/$31.8$ & $+.053$/$.183$/$30.0$ & $+.051$/$.182$/$29.9$ \\
NinaPro DB5 & $41.8$ & $+.092$/$.072$/$17.3$ & $+.091$/$.076$/$18.1$ & $+.085$/$.078$/$18.7$ & $+.039$/$.051$/$12.2$ & $+.038$/$.052$/$12.3$ \\
PTB-XL & $77.1$ & $+.165$/$.150$/$19.4$ & $+.159$/$.173$/$22.4$ & $+.151$/$.167$/$21.6$ & $+.133$/$.181$/$23.5$ & $+.135$/$.178$/$23.1$ \\
DrugBAN & $62.5$ & $+.084$/$.150$/$23.9$ & $+.082$/$.201$/$32.1$ & $+.071$/$.168$/$26.8$ & $+.037$/$.124$/$19.9$ & $+.051$/$.164$/$26.2$ \\
\bottomrule
\end{tabular}
\end{table}

\subsection{The corrector}
\label{sec:corrector}

\S\ref{sec:recal} treats the corrector as a plug-in. Table~\ref{tab:corrector} holds the frozen model and the splits fixed and replaces the neighbor average by a probe trained on the same $D_{\mathrm{source}}$. For the hard-label target, the probe is a multinomial logistic model on pool labels; for the cross-mask target, it is a ridge regression onto the richer mask's outputs. Each selects its regularization on $D_{\mathrm{fit}}$. Fitting and calibration are re-run for each corrector, since a threshold calibrated for one corrector does not transfer to another. Neither corrector dominates: the probe moves accuracy gain by at most $2.4$ points in either direction. Both stay inside the budget on every benchmark, and removing the gate puts both over it on four of the five. \textsc{Certify} is therefore not tied to the estimator it certifies.

\begin{table}[!htbp]
\centering
\footnotesize
\caption{\textbf{The same certificate over two correctors.} Gated gain, joint harm and ungated joint harm at $\alpha=0.2$, for the retrieval plug-in of \S\ref{sec:recal} and for a probe trained on the same $D_{\mathrm{source}}$. Everything but the corrector is held fixed; PTB-XL uses per-label accuracy. Bold marks the better of the two; $^{\ast}$ marks harm above the budget.}
\label{tab:corrector}
\begin{tabular}{lcccccc}
\toprule
& \multicolumn{2}{c}{Gated gain} & \multicolumn{2}{c}{Joint harm} & \multicolumn{2}{c}{Ungated harm} \\
\cmidrule(lr){2-3}\cmidrule(lr){4-5}\cmidrule(lr){6-7}
Benchmark & Retrieval & Probe & Retrieval & Probe & Retrieval & Probe \\
\midrule
IEMOCAP & $+0.008$ & $\bm{+0.017}$ & $\bm{0.086}$ & $0.103$ & $0.315^{\ast}$ & $0.458^{\ast}$ \\
AVE & $+0.084$ & $\bm{+0.108}$ & $\bm{0.178}$ & $0.184$ & $0.300^{\ast}$ & $0.285^{\ast}$ \\
NinaPro DB5 & $+0.092$ & $\bm{+0.093}$ & $\bm{0.072}$ & $0.081$ & $0.125$ & $0.118$ \\
PTB-XL & $+0.165$ & $\bm{+0.172}$ & $0.150$ & $\bm{0.140}$ & $0.239^{\ast}$ & $0.215^{\ast}$ \\
DrugBAN & $\bm{+0.084}$ & $+0.067$ & $\bm{0.150}$ & $0.153$ & $0.275^{\ast}$ & $0.359^{\ast}$ \\
\bottomrule
\end{tabular}
\end{table}

\paragraph{Latency and memory.}
\label{app:efficiency}
A probe is cheaper to run. Table~\ref{tab:efficiency} times the correction step on saved frozen representations. The timed path covers standardization, target construction, blending, scoring and gating; it excludes feature extraction, fitting and calibration. Both correctors use fixed settings: hard-label targets in a standardized space, uniform $k=20$ retrieval or a linear probe with regularization $C=1$, and unit temperature. Timings use float64 on one pinned Intel Xeon Gold 6244 CPU thread, with $200$ single-query calls after ten warm-up calls. Exact retrieval stores $O(N(d+K))$ values and scans all $N$ pool rows per query, whereas a linear probe needs $O(dK)$ coefficients. The probe is $14$--$31$ times faster for single queries here. Caching pool quantities or an approximate nearest-neighbor index would change this comparison; we have not measured either.

\begin{table}[!htbp]
\caption{\textbf{Correction overhead at fixed settings.} One condition per benchmark: audio-only AVE and IEMOCAP, the six limb leads on PTB-XL, and four-electrode NinaPro. $N$ is the retrieval-pool size, $d$ the feature dimension and $K$ the output dimension. Latency is the single-query median/$95$th percentile in milliseconds. State is the numerical arrays retained by the corrector, transform and scorer, in MiB, excluding the frozen model and the runtime.}
\label{tab:efficiency}
\centering\footnotesize
\setlength{\tabcolsep}{3pt}
\begin{tabular}{@{}lrcccccc@{}}
\toprule
& & & & \multicolumn{2}{c}{Latency (ms)} & \multicolumn{2}{c}{State (MiB)} \\
\cmidrule(lr){5-6}\cmidrule(lr){7-8}
Benchmark & $N$ & $d$ & $K$ & Retrieval & Probe & Retrieval & Probe \\
\midrule
AVE & $33280$ & $64$ & $29$ & $6.786/6.985$ & $0.398/0.430$ & $23.614$ & $0.016$ \\
PTB-XL & $17084$ & $256$ & $5$ & $17.517/17.900$ & $0.898/0.938$ & $34.023$ & $0.666$ \\
IEMOCAP & $4446$ & $768$ & $4$ & $11.195/11.540$ & $0.366/0.397$ & $26.198$ & $0.035$ \\
NinaPro & $14259$ & $128$ & $41$ & $5.544/5.768$ & $0.403/0.424$ & $18.387$ & $0.043$ \\
\bottomrule
\end{tabular}
\end{table}

\subsection{The action score}
\label{sec:actionscore}

Proposition~\ref{thm:noharm} holds for any score fixed before $D_{\mathrm{conf}}$ is read, so the choice of score can move gated gain and the intervention rate but not the bound. Table~\ref{tab:actionscore} holds everything else fixed and varies only the map from the five features of App.~\ref{app:erm} to a score. Each learner keeps its library defaults apart from one size knob.

The score family has limited effect: no learner beats the deployed logistic model by more than $0.002$ gain anywhere. The random forest is the most conservative variant on every benchmark and gives up a third of the gain on PTB-XL. The feature blocks show that the features are redundant with each other rather than individually useless. Removing any single feature costs little, with one exception, AVE without the $\ell_1$ distance, yet no single feature reproduces the full score everywhere. The agreement flag alone illustrates the degenerate case of App.~\ref{app:erm}. On AVE and NinaPro it is identically one, so the score is constant and conformal risk control can only admit everything or nothing. It admits nothing on AVE and everything on NinaPro, and the budget holds in both cases.

\begin{table}[!htbp]
\centering\scriptsize
\caption{\textbf{Learner and features of the action score.} Gated gain and joint harm at $\alpha=0.2$, written gain/harm. Everything but the score is held fixed. Evaluation cells: $6$ on AVE, $80$ on NinaPro, $18$ on PTB-XL, $175$ on IEMOCAP and $27$ on DrugBAN.}
\label{tab:actionscore}
\begin{tabular}{@{}lccccc@{}}
\toprule
Action score & AVE & NinaPro & PTB-XL & IEMOCAP & DrugBAN \\
\midrule
\multicolumn{6}{@{}l}{\emph{Score family, all five features}}\\
logistic (deployed)              & $+0.084$/$0.18$ & $+0.092$/$0.07$ & $+0.165$/$0.15$ & $+0.008$/$0.09$ & $+0.084$/$0.15$ \\
decision tree, depth $3$         & $+0.083$/$0.17$ & $+0.092$/$0.08$ & $+0.162$/$0.15$ & $+0.006$/$0.09$ & $+0.079$/$0.12$ \\
random forest, $200$ trees       & $+0.072$/$0.11$ & $+0.086$/$0.04$ & $+0.112$/$0.08$ & $+0.007$/$0.07$ & $+0.067$/$0.06$ \\
$k$NN, $k=25$                    & $+0.083$/$0.18$ & $+0.092$/$0.08$ & $+0.163$/$0.15$ & $+0.008$/$0.09$ & $+0.084$/$0.15$ \\
MLP, one hidden layer of $16$    & $+0.084$/$0.17$ & $+0.092$/$0.07$ & $+0.164$/$0.15$ & $+0.009$/$0.08$ & $+0.084$/$0.15$ \\
\midrule
\multicolumn{6}{@{}l}{\emph{Logistic score, one feature removed}}\\
without base confidence          & $+0.084$/$0.17$ & $+0.091$/$0.07$ & $+0.162$/$0.16$ & $+0.009$/$0.09$ & $+0.084$/$0.15$ \\
without base entropy             & $+0.084$/$0.18$ & $+0.092$/$0.07$ & $+0.165$/$0.15$ & $+0.009$/$0.09$ & $+0.084$/$0.16$ \\
without target confidence        & $+0.084$/$0.18$ & $+0.091$/$0.07$ & $+0.164$/$0.15$ & $+0.008$/$0.09$ & $+0.083$/$0.16$ \\
without agreement flag           & $+0.084$/$0.18$ & $+0.092$/$0.07$ & $+0.165$/$0.15$ & $+0.009$/$0.09$ & $+0.084$/$0.15$ \\
without $\ell_1$ distance        & $+0.050$/$0.18$ & $+0.089$/$0.10$ & $+0.165$/$0.16$ & $+0.009$/$0.09$ & $+0.082$/$0.16$ \\
\midrule
\multicolumn{6}{@{}l}{\emph{Logistic score, one feature only}}\\
base confidence only             & $+0.071$/$0.18$ & $+0.090$/$0.11$ & $+0.165$/$0.16$ & $+0.007$/$0.10$ & $+0.070$/$0.16$ \\
base entropy only                & $+0.071$/$0.18$ & $+0.090$/$0.11$ & $+0.158$/$0.17$ & $+0.008$/$0.10$ & $+0.070$/$0.16$ \\
target confidence only           & $+0.039$/$0.18$ & $+0.087$/$0.10$ & $+0.158$/$0.18$ & $+0.005$/$0.11$ & $+0.070$/$0.16$ \\
agreement flag only              & $+0.000$/$0.00$ & $+0.092$/$0.12$ & $+0.080$/$0.06$ & $+0.007$/$0.08$ & $+0.018$/$0.06$ \\
$\ell_1$ distance only           & $+0.084$/$0.18$ & $+0.091$/$0.07$ & $+0.153$/$0.16$ & $+0.008$/$0.08$ & $+0.068$/$0.14$ \\
\bottomrule
\end{tabular}
\end{table}

\subsection{The retrieval pool and the calibration set}
\label{app:pool}

Table~\ref{tab:source-pool} varies one offline role at a time and keeps the others fixed. On CMU-MOSEI, $D_{\mathrm{source}}$ is carved from deployment data that the models never trained on. There the unlabeled cross-mask target keeps more gain than hard labels at every pool size. Flipping pool labels removes the hard-label gain, while GUARD switches to the cross-mask target on $D_{\mathrm{fit}}$ and keeps its own. A smaller pool lowers gain and joint harm together, because the gate declines more often. On DrugBAN, hard-label gain grows steadily with pool size. Shrinking $D_{\mathrm{conf}}$ to $50$ points keeps mean joint harm inside the budget and barely moves gain. Under the cluster shift, the origin of the pool matters more than its size: $1.2$k--$2.4$k deployment rows beat the $10$k--$12$k-row source pool, and adding the source pool to them does not help.

On the DrugBAN random splits of Table~\ref{tab:crossmask}, $D_{\mathrm{source}}$ is a subsample of the model's training data, on which the all-inputs model is nearly exact. When the pool is instead carved from deployment data that the model never trained on, cross-mask transfer still matches or exceeds hard-label gain: $+0.007$ against $+0.004$ on Human, $+0.075$ against $+0.071$ on BindingDB, and $+0.077$ against $+0.076$ on BioSNAP.

\begin{table}[!htbp]
\caption{\textbf{Varying one offline role at a time}, at $\alpha=0.2$. (a) CMU-MOSEI: three frozen models, modality subsets $A$, $V$ and $AV$, five splits; accuracy gain on non-neutral samples as $D_{\mathrm{source}}$ shrinks, and as a fraction of its labels is flipped at full size. (b) DrugBAN random splits, three datasets, three seeds and the two conditions of Table~\ref{tab:crossmask}: hard-label AUROC gain as the source pool shrinks, three subsamples per size. (c) Gain and mean joint harm as $D_{\mathrm{conf}}$ shrinks, over $8$ (MOSEI) or $20$ (DrugBAN, seed $42$) subsamples per size. (d) DrugBAN cluster splits: AUROC gain of GUARD by pool origin; \emph{matched} subsamples the source pool to the size of the deployment pool, \emph{source} uses all of it, and \emph{both} adds it to the deployment pool.}
\label{tab:source-pool}
\centering\footnotesize
\setlength{\tabcolsep}{6pt}
\begin{tabular}{@{}lccccc@{}}
\toprule
\multicolumn{6}{@{}l}{\emph{(a) CMU-MOSEI}}\\
$|D_{\mathrm{source}}|$ & $50$ & $150$ & $400$ & full & \\
\midrule
Hard-label gain  & $+0.001$ & $-0.003$ & $+0.004$ & $+0.015$ & \\
Cross-mask gain  & $+0.001$ & $+0.004$ & $+0.014$ & $+0.035$ & \\
GUARD gain       & $+0.000$ & $+0.002$ & $+0.013$ & $+0.035$ & \\
GUARD joint harm & $0.025$ & $0.041$ & $0.081$ & $0.094$ & \\
\midrule
Flipped pool labels & $0$ & $0.1$ & $0.2$ & $0.3$ & $0.4$ \\
\midrule
Hard-label gain  & $+0.015$ & $+0.008$ & $+0.005$ & $+0.003$ & $-0.000$ \\
GUARD gain       & $+0.035$ & $+0.035$ & $+0.034$ & $+0.035$ & $+0.035$ \\
GUARD picks cross-mask (\%) & $100$ & $96$ & $93$ & $98$ & $98$ \\
\midrule
\multicolumn{6}{@{}l}{\emph{(b) DrugBAN, random}}\\
$|D_{\mathrm{source}}|$ & $100$ & $300$ & $1000$ & $3000$ & full \\
\midrule
Hard-label AUROC gain & $+0.008$ & $+0.016$ & $+0.033$ & $+0.053$ & $+0.079$ \\
Joint harm       & $0.144$ & $0.148$ & $0.145$ & $0.131$ & $0.128$ \\
\midrule
\multicolumn{6}{@{}l}{\emph{(c) Calibration size}}\\
$|D_{\mathrm{conf}}|$ & $50$ & $100$ & $200$ & full & \\
\midrule
CMU-MOSEI gain       & $+0.033$ & $+0.035$ & $+0.035$ & $+0.035$ & \\
CMU-MOSEI joint harm & $0.086$ & $0.091$ & $0.093$ & $0.094$ & \\
DrugBAN AUROC gain   & $+0.077$ & $+0.076$ & $+0.076$ & $+0.076$ & \\
DrugBAN joint harm   & $0.136$ & $0.138$ & $0.132$ & $0.133$ & \\
\midrule
\multicolumn{6}{@{}l}{\emph{(d) DrugBAN, cluster}}\\
Pool & deployment & matched & source & both & \\
\midrule
BioSNAP   & $+0.091$ & $+0.042$ & $+0.048$ & $+0.086$ & \\
BindingDB & $+0.307$ & $+0.043$ & $+0.032$ & $+0.303$ & \\
\bottomrule
\end{tabular}
\end{table}

\subsection{Label budgets}
\label{app:budget}

Role allocations were fixed before any run. Each arm selects its settings on $D_{\mathrm{fit}}$ exactly as GUARD does, and smaller budgets are nested in larger ones. Table~\ref{tab:budget-full} adds joint harm and the Learn-then-Test gain to Table~\ref{tab:budget}, over five deployment partitions of CMU-MOSEI and three seeds each of OPPORTUNITY and BioSNAP cluster. Table~\ref{tab:budget-pool} covers benchmarks whose pool already holds training labels, so the budget covers only $D_{\mathrm{fit}}$ and $D_{\mathrm{conf}}$. Across the $102$ budget combinations of the two tables, the mean joint harm of GUARD stays below $\alpha$.

\begin{table}[!htbp]
\caption{\textbf{Label budget in full.} Cells give gain/joint harm under GUARD, then the Learn-then-Test gain in brackets, at $\alpha=0.2$. Hard arms use the full budget across three roles; cross arms use it across two, with the whole pool unlabeled.}
\label{tab:budget-full}
\centering\scriptsize
\setlength{\tabcolsep}{3pt}
\begin{tabular}{@{}llcccc@{}}
\toprule
 & & \multicolumn{2}{c}{Retrieval} & \multicolumn{2}{c}{Probe} \\
\cmidrule(lr){3-4}\cmidrule(lr){5-6}
Benchmark & Labels & hard & cross & hard & cross \\
\midrule
CMU-MOSEI & $60$ & $-.012$/$.065$ ($-.003$) & $+.017$/$.032$ ($+.014$) & $-.003$/$.025$ ($-.001$) & $+.020$/$.062$ ($+.011$) \\
 & $120$ & $-.007$/$.039$ ($-.005$) & $+.027$/$.047$ ($+.022$) & $-.007$/$.051$ ($-.001$) & $+.006$/$.032$ ($+.008$) \\
 & $240$ & $-.003$/$.031$ ($-.003$) & $+.025$/$.050$ ($+.020$) & $-.003$/$.048$ ($-.001$) & $+.012$/$.074$ ($+.008$) \\
 & $480$ & $+.002$/$.022$ ($+.001$) & $+.036$/$.067$ ($+.029$) & $+.001$/$.023$ ($+.000$) & $+.022$/$.079$ ($+.016$) \\
 & $960$ & $-.001$/$.055$ ($-.001$) & $+.045$/$.090$ ($+.043$) & $+.002$/$.032$ ($+.001$) & $+.023$/$.064$ ($+.021$) \\
 & $1800$ & $+.012$/$.084$ ($+.009$) & $+.041$/$.078$ ($+.039$) & $+.007$/$.056$ ($+.007$) & $+.025$/$.082$ ($+.023$) \\
 & full & $+.029$/$.102$ ($+.029$) & $+.050$/$.104$ ($+.049$) & $+.015$/$.082$ ($+.014$) & $+.026$/$.090$ ($+.024$) \\
OPPORTUNITY & $60$ & $+.028$/$.104$ ($+.019$) & $+.021$/$.047$ ($+.016$) & $+.027$/$.132$ ($+.021$) & $+.021$/$.071$ ($+.016$) \\
 & $120$ & $+.029$/$.122$ ($+.023$) & $+.026$/$.082$ ($+.020$) & $+.034$/$.081$ ($+.032$) & $+.030$/$.086$ ($+.025$) \\
 & $240$ & $+.035$/$.125$ ($+.031$) & $+.032$/$.087$ ($+.030$) & $+.052$/$.136$ ($+.046$) & $+.042$/$.094$ ($+.039$) \\
 & $480$ & $+.061$/$.122$ ($+.060$) & $+.058$/$.110$ ($+.045$) & $+.058$/$.145$ ($+.053$) & $+.048$/$.113$ ($+.042$) \\
 & $960$ & $+.128$/$.139$ ($+.123$) & $+.048$/$.117$ ($+.047$) & $+.101$/$.156$ ($+.092$) & $+.050$/$.137$ ($+.049$) \\
 & $1800$ & $+.147$/$.157$ ($+.141$) & $+.054$/$.143$ ($+.046$) & $+.115$/$.161$ ($+.106$) & $+.045$/$.145$ ($+.044$) \\
 & full & $+.178$/$.170$ ($+.176$) & $+.056$/$.154$ ($+.053$) & $+.135$/$.172$ ($+.133$) & $+.050$/$.164$ ($+.049$) \\
BioSNAP cluster & $60$ & $+.018$/$.104$ ($-.001$) & $+.038$/$.143$ ($+.021$) & $+.020$/$.080$ ($-.002$) & $-.006$/$.086$ ($-.005$) \\
 & $120$ & $+.029$/$.124$ ($+.025$) & $+.044$/$.100$ ($+.017$) & $+.014$/$.031$ ($+.001$) & $+.007$/$.126$ ($+.000$) \\
 & $240$ & $+.015$/$.116$ ($+.006$) & $+.051$/$.105$ ($+.049$) & $+.008$/$.080$ ($+.011$) & $+.006$/$.129$ ($+.001$) \\
 & $480$ & $+.043$/$.085$ ($+.036$) & $+.050$/$.127$ ($+.050$) & $+.032$/$.162$ ($+.018$) & $+.014$/$.179$ ($+.006$) \\
 & $960$ & $+.058$/$.160$ ($+.058$) & $+.051$/$.188$ ($+.053$) & $+.067$/$.154$ ($+.059$) & $+.021$/$.179$ ($+.016$) \\
 & $1800$ & $+.079$/$.124$ ($+.075$) & $+.055$/$.161$ ($+.054$) & $+.098$/$.149$ ($+.094$) & $+.026$/$.191$ ($+.024$) \\
 & full & $+.101$/$.140$ ($+.097$) & $+.057$/$.161$ ($+.054$) & $+.131$/$.131$ ($+.127$) & $+.026$/$.199$ ($+.024$) \\
\bottomrule
\end{tabular}
\end{table}

\begin{table}[!htbp]
\caption{\textbf{Deployment labels when pool labels already exist.} Gain, joint harm and apply rate under GUARD and under Learn-then-Test with the same tightening, with half the budget on $D_{\mathrm{fit}}$ and half on $D_{\mathrm{conf}}$. IEMOCAP uses six masks, five folds and one cut, so its \emph{full} row differs from Table~\ref{tab:allbench}.}
\label{tab:budget-pool}
\centering\scriptsize
\setlength{\tabcolsep}{7pt}
\begin{tabular}{@{}llcccccc@{}}
\toprule
 & & \multicolumn{3}{c}{GUARD} & \multicolumn{3}{c}{LTT} \\
\cmidrule(lr){3-5}\cmidrule(lr){6-8}
Benchmark & Labels & gain & harm & apply (\%) & gain & harm & apply (\%) \\
\midrule
PTB-XL & $40$ & $+.128$ & $.125$ & $58$ & $+.056$ & $.045$ & $22$ \\
 & $80$ & $+.131$ & $.114$ & $57$ & $+.075$ & $.048$ & $28$ \\
 & $160$ & $+.141$ & $.115$ & $62$ & $+.103$ & $.069$ & $40$ \\
 & $320$ & $+.152$ & $.128$ & $67$ & $+.135$ & $.095$ & $54$ \\
 & $640$ & $+.158$ & $.139$ & $72$ & $+.150$ & $.116$ & $63$ \\
 & full & $+.165$ & $.150$ & $77$ & $+.161$ & $.138$ & $73$ \\
AVE & $40$ & $+.054$ & $.094$ & $34$ & $+.024$ & $.031$ & $14$ \\
 & $80$ & $+.068$ & $.122$ & $38$ & $+.040$ & $.054$ & $20$ \\
 & $160$ & $+.068$ & $.177$ & $54$ & $+.051$ & $.067$ & $22$ \\
 & $320$ & $+.086$ & $.171$ & $54$ & $+.083$ & $.118$ & $41$ \\
 & $640$ & $+.087$ & $.160$ & $55$ & $+.087$ & $.122$ & $46$ \\
 & full & $+.084$ & $.178$ & $61$ & $+.084$ & $.154$ & $56$ \\
IEMOCAP & $40$ & $+.010$ & $.062$ & $18$ & $+.004$ & $.019$ & $9$ \\
 & $80$ & $+.013$ & $.038$ & $13$ & $+.009$ & $.017$ & $8$ \\
 & $160$ & $+.011$ & $.053$ & $20$ & $+.010$ & $.028$ & $13$ \\
 & $320$ & $+.014$ & $.067$ & $25$ & $+.012$ & $.049$ & $18$ \\
 & $640$ & $+.019$ & $.078$ & $26$ & $+.016$ & $.062$ & $21$ \\
 & full & $+.013$ & $.080$ & $27$ & $+.010$ & $.062$ & $23$ \\
\bottomrule
\end{tabular}
\end{table}

\subsection{Changing deployment conditions}
\label{app:shift}

Proposition~\ref{thm:noharm} assumes that calibration and deployment samples are exchangeable. Two deliberate violations show what happens without it. Calibrating on one subject and deploying on another makes the gate more conservative: OPPORTUNITY's intervention rate falls from $0.650$ to $0.615$, and mean joint harm from $0.169$ to $0.147$. Calibrating at a mild degradation and deploying at a severe one is the damaging direction, since calibration then looks easier than deployment. PTB-XL's rate rises from $0.739$ to $0.844$ and mean joint harm from $0.141$ to $0.144$, still inside the budget.

Table~\ref{tab:mixed} lets every sample draw its own availability pattern, with one corrector and either one threshold or one threshold per mask (Cor.~\ref{cor:mask}). The per-pattern harm matters when the mix of patterns drifts, since a new mix only reweights those harms. Per-mask tightening uses few fit samples and gives up more gain than the pooled step, so per-mask calibration works best without it.

Table~\ref{tab:reuse} takes parts of the adapter from another condition on the same samples. Taking only $D_{\mathrm{conf}}$ from another condition keeps mean harm inside the budget here, and gain falls by up to half on IEMOCAP; this case lies outside Proposition~\ref{thm:noharm}. The whole adapter does not transfer. Gain turns negative on CMU-MOSEI and IEMOCAP in both directions, and on NinaPro from few electrodes to many. The gate still limits the damage: the same candidate applied everywhere reaches harm $0.43$, $0.42$ and $0.66$ on these three benchmarks.

\begin{table}[!htbp]
\caption{\textbf{Mixed availability.} Gain, pooled joint harm and the largest per-pattern mean joint harm (\emph{worst}), at $\alpha=0.2$. Patterns with fewer than $20$ test samples are skipped. On IEMOCAP, every modality is absent independently with probability $\eta$, and a sample that would lose all three keeps one.}
\label{tab:mixed}
\centering\scriptsize
\setlength{\tabcolsep}{2.4pt}
\begin{tabular}{@{}lcccccccccccc@{}}
\toprule
 & \multicolumn{3}{c}{One threshold} & \multicolumn{3}{c}{One threshold, tightened} & \multicolumn{3}{c}{Per mask} & \multicolumn{3}{c}{Per mask, tightened} \\
\cmidrule(lr){2-4}\cmidrule(lr){5-7}\cmidrule(lr){8-10}\cmidrule(lr){11-13}
Benchmark & gain & harm & worst & gain & harm & worst & gain & harm & worst & gain & harm & worst \\
\midrule
CMU-MOSEI & $+0.033$ & $0.194$ & $0.240$ & $+0.023$ & $0.076$ & $0.107$ & $+0.033$ & $0.193$ & $0.207$ & $+0.015$ & $0.037$ & $0.055$ \\
NinaPro DB5 & $+0.076$ & $0.140$ & $0.198$ & $+0.075$ & $0.076$ & $0.119$ & $+0.076$ & $0.134$ & $0.176$ & $+0.073$ & $0.054$ & $0.063$ \\
IEMOCAP, $\eta=0.3$ & $+0.005$ & $0.167$ & $0.293$ & $+0.004$ & $0.069$ & $0.186$ & $+0.006$ & $0.158$ & $0.183$ & $+0.002$ & $0.053$ & $0.064$ \\
IEMOCAP, $\eta=0.5$ & $+0.010$ & $0.182$ & $0.269$ & $+0.013$ & $0.079$ & $0.133$ & $+0.010$ & $0.176$ & $0.186$ & $+0.009$ & $0.050$ & $0.066$ \\
\bottomrule
\end{tabular}
\end{table}

\begin{table}[!htbp]
\caption{\textbf{Reusing an adapter at another condition.} Mean gain and joint harm of GUARD at $\alpha=0.2$. \emph{Calibration only}: corrector and score are fitted at the deployed condition, and $D_{\mathrm{conf}}$ comes from the other one. \emph{Whole adapter}: corrector, score and threshold all come from the other condition.}
\label{tab:reuse}
\centering\footnotesize
\setlength{\tabcolsep}{8pt}
\begin{tabular}{@{}llcccc@{}}
\toprule
 & & \multicolumn{2}{c}{Calibration only} & \multicolumn{2}{c}{Whole adapter} \\
\cmidrule(lr){3-4}\cmidrule(lr){5-6}
Benchmark & Source to target & gain & harm & gain & harm \\
\midrule
CMU-MOSEI & same condition & $+0.050$ & $0.104$ & $+0.050$ & $0.104$ \\
 & richer to poorer & $+0.032$ & $0.057$ & $-0.058$ & $0.227$ \\
 & poorer to richer & $+0.036$ & $0.059$ & $-0.022$ & $0.186$ \\
IEMOCAP & same condition & $+0.013$ & $0.080$ & $+0.013$ & $0.080$ \\
 & richer to poorer & $+0.012$ & $0.076$ & $-0.020$ & $0.218$ \\
 & poorer to richer & $+0.007$ & $0.042$ & $-0.036$ & $0.125$ \\
NinaPro DB5 & same condition & $+0.088$ & $0.071$ & $+0.088$ & $0.071$ \\
 & richer to poorer & $+0.074$ & $0.030$ & $+0.016$ & $0.055$ \\
 & poorer to richer & $+0.084$ & $0.072$ & $-0.083$ & $0.520$ \\
\bottomrule
\end{tabular}
\end{table}

\subsection{Harm budget and tolerance}
\label{app:delta}

\begin{figure}[!htbp]
\centering
\includegraphics[width=\linewidth]{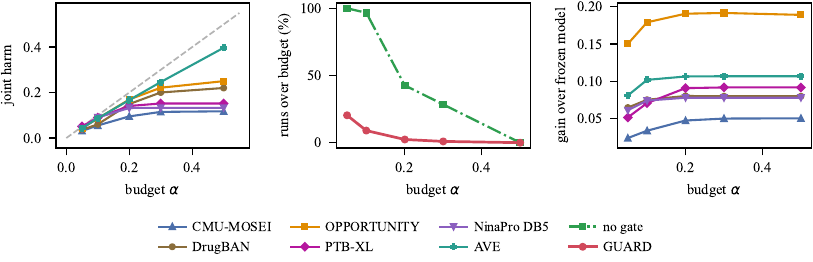}
\caption{Risk and utility under different budgets $\alpha$, over $1055$ runs spanning every
benchmark but IEMOCAP, which is not swept. \emph{Left:} empirical joint harm against the budget, with the dashed diagonal at
$\alpha$; every per-benchmark mean stays below it. \emph{Middle:} the percentage of runs over
budget, for GUARD and for the same corrector ungated. \emph{Right:} gated gain in each benchmark's
own metric, which rises with the budget and levels off by $\alpha=0.3$. Target, neighborhood and blend weight are
selected once on $D_{\mathrm{fit}}$ and held fixed, so $\alpha$ alone varies and the
ungated corrector is the same at every budget. Intervention rates are in Figure~\ref{fig:apply}.}
\label{fig:alpha}
\end{figure}

\begin{figure}[!htbp]
\centering
\includegraphics[width=0.62\linewidth]{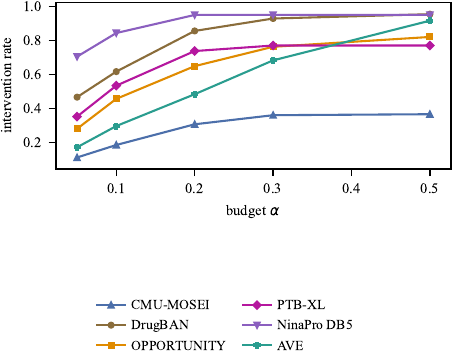}
\caption{Intervention rate against the budget. Before tightening, a larger budget can only lower the CRC threshold for a fixed scorer and calibration sample. The plotted policy also applies fit-split tightening, so its intervention rate need not be monotone and may saturate below one.}
\label{fig:apply}
\end{figure}

Figure~\ref{fig:alpha} sweeps the budget $\alpha$. Target, neighborhood and blend weight are selected once on $D_{\mathrm{fit}}$, so only $\alpha$ varies. Mean joint harm stays below the budget on every benchmark, and gated gain rises with the budget and levels off by $\alpha=0.3$. Figure~\ref{fig:apply} gives the intervention rate. Table~\ref{tab:delta} sweeps the tolerance $\delta$ with the retrieval settings fixed, so only the harm event moves. Gain and joint harm barely change between $\delta=0$, where any loss increase counts as harm, and $\delta=0.2$. Applied corrections turn a wrong frozen decision correct about $1.7$ times as often as the reverse on CMU-MOSEI and $2.6$ times as often on DrugBAN.

\begin{table}[!htbp]
\caption{\textbf{Harm tolerance $\delta$} at $\alpha=0.2$. CMU-MOSEI: three frozen models, modality subsets $A$, $V$ and $AV$, five splits; gain in accuracy on non-neutral samples. DrugBAN: the settings of Table~\ref{tab:crossmask} under the two conditions used there; gain in AUROC. \emph{Flip}$^-$ and \emph{flip}$^+$ are the shares of test samples on which an applied correction turns a correct frozen decision wrong, or a wrong one correct.}
\label{tab:delta}
\centering\footnotesize
\setlength{\tabcolsep}{4pt}
\begin{tabular}{@{}lcccccccc@{}}
\toprule
& \multicolumn{4}{c}{CMU-MOSEI} & \multicolumn{4}{c}{DrugBAN} \\
\cmidrule(lr){2-5}\cmidrule(lr){6-9}
$\delta$ & Gain & Joint harm & Flip$^-$ & Flip$^+$ & Gain & Joint harm & Flip$^-$ & Flip$^+$ \\
\midrule
$0$ & $+0.034$ & $0.093$ & $0.034$ & $0.059$ & $+0.099$ & $0.135$ & $0.070$ & $0.187$ \\
$0.02$ & $+0.035$ & $0.097$ & $0.036$ & $0.062$ & $+0.099$ & $0.134$ & $0.071$ & $0.188$ \\
$0.05$ & $+0.035$ & $0.094$ & $0.037$ & $0.063$ & $+0.100$ & $0.133$ & $0.072$ & $0.189$ \\
$0.1$ & $+0.036$ & $0.087$ & $0.040$ & $0.067$ & $+0.101$ & $0.134$ & $0.074$ & $0.192$ \\
$0.2$ & $+0.037$ & $0.059$ & $0.043$ & $0.071$ & $+0.105$ & $0.136$ & $0.078$ & $0.196$ \\
\bottomrule
\end{tabular}
\end{table}

\section{Controlled Separation of Error, Ambiguity and Gap}
\label{app:synthetic}

On real models, the claim that correction gain tracks the Bayes gap rather than raw error is correlational because gap, ambiguity, and error cannot be manipulated independently. We therefore construct a synthetic family in which they can be controlled: $X\sim\mathcal N(0,I_{10})$, $\pi(X)=\mathrm{softmax}(WX/\tau)$, $Y\sim\mathrm{Cat}(\pi(X))$, and model prediction $m(X)=\mathrm{softmax}(WX/\tau+\gamma VX)$. The temperature $\tau$ controls ambiguity ($\operatorname{tr}\Cov(e_Y\mid X)$), while $\gamma$ controls systematic conditional bias. We tune three configurations to have matched base accuracy ($\approx0.63$--$0.66$) but different sources of error. The full GUARD pipeline (a source-$k$NN corrector and $\widehat\beta$ fitted on held-out data) then runs identically in each configuration over three seeds.

\begin{table}[ht]
\centering
\footnotesize
\caption{\textbf{Three matched-accuracy configurations.} A separate run of the family behind Figure~\ref{fig:synthetic}, with configurations of matched base accuracy whose error comes from the Bayes gap in different proportions. $\PH_{\mathrm{true}}=\E\|d(X)\|_2^2$ is potential headroom, the mean squared Bayes gap; \emph{Loss gain} is $L_{\mathrm{base}}-L_{\mathrm{corrected}}$. Mean over three seeds; the family is specified in App.~\ref{app:synthetic}.}
\label{tab:app-synthetic}
\begin{tabular}{lccccc}
\toprule
Configuration & Base acc & $\PH_{\mathrm{true}}$ & $\widehat\beta$ & Loss gain & $\Delta$acc \\
\midrule
Bias-dominant & $0.64$ & $0.39$ & $0.75$ & $\bm{+1.02}$ & $+0.11$ \\
Mixed & $0.65$ & $0.23$ & $0.62$ & $+0.37$ & $+0.05$ \\
Ambiguity-dominant & $0.65$ & $0.00$ & $0.03$ & $-0.00$ & $+0.00$ \\
\bottomrule
\end{tabular}
\end{table}

Figure~\ref{fig:synthetic} sweeps this family densely: for each $\tau$ on a grid, $\gamma$ is bisected until the base accuracy returns to the target, so accuracy is matched by construction at every point rather than at three tuned ones. Over twelve points and ten seeds the base accuracy is $0.650$ with a standard deviation of $0.010$, while potential headroom runs from $0.369$ to $0.0005$. Across the twelve points, accuracy gain and $\PH_{\mathrm{true}}$ correlate at $0.990$, with a $95\%$ interval of $[0.980,0.993]$ from resampling seeds; over the $120$ individual runs the correlation is $0.962$. Two predictions hold along the sweep, and Table~\ref{tab:app-synthetic} adds three configurations from a separate three-seed run. First, at fixed raw error, gain increases with true headroom and vanishes for the Bayes-optimal model even though its accuracy is no better. Second, the fitted $\widehat\beta$ falls from $0.75$ to $0.03$ as the gap disappears, although the fitting data do not explicitly mark the absence of headroom.

Table~\ref{tab:app-synthetic} makes the control explicit at matched base accuracy: the ambiguity-dominated configuration gains nothing on either scale, while the bias-dominated one gains $+0.11$ in accuracy and $+1.02$ in loss. This is the opposite of what the heuristic ``correction helps weak models'' would predict.

\section{Scope and Limitations}
\label{sec:limitations}
This appendix specifies the domain over which each result of the paper applies.

\paragraph{Frozen-model scope.}
Theorem~\ref{thm:reduction} is exact for the information supplied to a corrector. If GUARD reads only $U=\phi(X)$, including the frozen prediction and retrieval features, its Bayes target is $\pi_U=\E[t(Y)\mid U]$. For square-integrable $t(Y)$ and a square-integrable $U$-measurable prediction $p$,
\begin{equation}
\label{eq:accessible-gap}
\E\|p-\pi(X)\|^2=\E\|p-\pi_U\|^2+\E\|\pi_U-\pi(X)\|^2.
\end{equation}
The cross term vanishes by conditioning on $U$. The second term is information lost by the accessible representation and cannot be removed by a $U$-measurable correction. Retrieval in that space need not estimate $\pi(X)$ itself. Feature-space reduction concerns a local quadratic surrogate; output-space containment is exact for measurable heads and unrestricted measurable output edits (Props.~\ref{prop:feature},~\ref{prop:output}). Containment is an expressivity statement, not an algorithm for computing a feature repair without access to the head.

\paragraph{Data and access requirements.}
GUARD reads the deployed model's predictions and representations and modifies neither. Beyond
that it assumes labeled data drawn from the deployment condition: $D_{\mathrm{source}}$ supplies
the retrieval pool, $D_{\mathrm{fit}}$ fits the blend weight and the action score, and
$D_{\mathrm{conf}}$ calibrates the threshold. The default hard-label target requires labels for the
pool, and a pool drawn from a distribution far from deployment reduces utility without affecting
validity. Cross-mask transfer removes labels from retrieval only; the accuracy screen, $\widehat\beta$-selection and \textsc{Certify} continue to use them. An adapter is fitted to one condition and does not transfer to another (App.~\ref{app:shift}), so a deployment with several availability patterns needs one per pattern. The mask is observed, so this choice can be made at query time.

\paragraph{Scope of the statistical guarantee.}
The present instantiation of \textsc{Certify} applies to classification. Nothing in conformal risk
control is specific to that case, but the action score and its features are defined over a
predicted distribution, so a regression instantiation would redefine both.
Proposition~\ref{thm:noharm} assumes the calibration split and the deployment sample are
exchangeable, and it controls joint marginal harm: it bounds the rate at which a harmful
correction is applied, not the safety of any single query and not the harm conditional on the
decision to apply. Distribution shift after calibration falls outside the formal guarantee;
App.~\ref{app:shift} reports two deliberate violations, and
App.~\ref{app:noharm} gives the formal statement and the mask-conditional variant.

\paragraph{Scope of the cross-mask result.}
Population recovery holds when $\E[d_T\mid X_S]=0$; recovery by finite-pool estimators additionally requires consistency. The observed sigma-fields, including mask context, must be nested. A conditionally correct richer predictor suffices for recovery, and Proposition~\ref{prop:transfer} uses this stronger condition with matched weights and blend. It is a convex-risk comparison before adaptive selection and gating, and need not hold after tempering the teacher. It does not imply dominance in accuracy or gated gain. Because conditional correctness is not identifiable from unlabeled paired data alone (App.~\ref{app:inherit}), no assumption-free unlabeled diagnostic can certify it in general. GUARD selects its target on labeled $D_{\mathrm{fit}}$; App.~\ref{app:screen} provides a descriptive accuracy screen. Realizing correction headroom additionally depends on the quality of the target estimator.

\paragraph{Sensitive deployments.}
Several of the datasets used here record people, and PTB-XL contains clinical ECGs. Proposition~\ref{thm:noharm} bounds a marginal rate over a deployment distribution: it certifies no individual prediction and does not by itself bound harm within a subgroup. GUARD changes the end-to-end output even when the backbone is frozen. Clinical use would therefore require validation of the corrected system and assessment under applicable regulatory change control; unchanged weights alone imply neither regulatory exemption nor clinical safety~\citep{fda2025pccp}.

\FloatBarrier
\section{Proofs for the Correction Reduction}
\subsection{The correction reduction}
\label{app:proof-reduction}

\paragraph{Loss conventions.}
\label{app:loss-conventions}
The gradient in \eqref{eq:canonical} is with respect to scores $z$, whereas blending and harm evaluation use predictions $p$. We distinguish $\ell(y,z)$ from $\bar\ell(y,p)$ throughout. For categorical cross-entropy,
\[
\ell(y,z)=\log\sum_c e^{z_c}-z_y,
\qquad m(z)=\softmax(z),\qquad \bar\ell(y,p)=-\log p_y.
\]
For squared error, $\ell(y,z)=\frac12\|z-y\|^2$ and $m(z)=z$. The implemented Brier score is $\bar\ell(y,p)=\frac12\|p-e_y\|^2$. It is convex in $p$; composing it with softmax does not give the canonical score gradient $p-e_y$. The unhalved Brier convention used for $\PH_{\mathrm{true}}$ and the excess-risk formulas below is twice this loss.

For multi-label outputs, summed binary cross-entropy has gradient $\operatorname{sigmoid}(z)-y$. The implementation averages over the $K$ labels, giving gradient $(\operatorname{sigmoid}(z)-y)/K$. This positive scale leaves the minimizer unchanged but sets the units of $\delta$: the reported multi-label tolerance is per label. The Bayes targets are marginal label probabilities; no conditional independence of the labels is required. Its excess risk is the mean of the coordinatewise Bernoulli KL divergences, rather than a categorical KL divergence between multi-label vectors.

The canonical identities and the convex-risk comparison use these ideal losses. For numerical evaluation, categorical probabilities inside logarithms are lower-clipped at $10^{-12}$, and Bernoulli probabilities are clipped to $[10^{-12},1-10^{-12}]$. The harm event and fit loss use the same implemented convention. CRC still applies to the resulting bounded indicator; the ideal gradient and convexity arguments are not claims about clipped losses at saturation.

\begin{proof}[Proof of Theorem~\ref{thm:reduction}]
On $\R^K$, \eqref{eq:canonical} implies a representation
\[
\ell(y,u)=A(u)-\langle t(y),u\rangle+c(y),\qquad \nabla A(u)=m(u).
\]
Indeed, for any fixed label $y_0$, take $A(u)=\ell(y_0,u)+\langle t(y_0),u\rangle$; the difference above has zero gradient and is constant in $u$. Integrability of $t(Y)$ and of the loss at one finite score ensures integrability of $c(Y)$. Thus the conditional risk has the explicit form
\[
R_X(u)=A(u)-\langle\pi(X),u\rangle+\E[c(Y)\mid X],
\qquad \nabla R_X(u)=m(u)-\pi(X).
\]
This proves \eqref{eq:exact-principle} without interchanging a derivative and an expectation. Every finite minimizer in the unconstrained score space has zero gradient. Convexity makes this condition sufficient. If $m(z(X))=\pi(X)$ almost surely, $R_X(z'(X))\ge R_X(z(X))$ pointwise for every measurable finite edit, and integration proves the risk claim whenever the risks are defined. Softmax reaches only the open simplex: an interior categorical target is attained by $z_c^\star=\log\pi_c+a$, with any scalar $a$; a boundary target is approached by a sequence of finite logits. Squared error attains its target at $z^\star=\pi(X)$.
\end{proof}

\paragraph{The geometry of headroom.}
For categorical log loss and half-squared loss respectively, the excess conditional risks are
\begin{equation}
\label{eq:loss-headroom}
R_X(p)-R_X(\pi)=\mathrm{KL}(\pi\|p),
\qquad
R_X(p)-R_X(\pi)=\tfrac12\|p-\pi\|^2,
\end{equation}
where $R_X$ here denotes prediction risk under $\bar\ell$. The first identity follows by subtracting the entropy of $\pi$; the second follows by expanding the square and using $\E[t(Y)-\pi\mid X]=0$. These are the classical Bregman geometries of a proper loss~\citep{banerjee2005bregman,gneiting2007proper}, two different geometries of the same target discrepancy. What Theorem~\ref{thm:reduction} adds to them is a statement about deployment: under a frozen predictor, the part of the divergence that any measurable edit can remove is fixed by the information the corrector observes, which is why degradation severity and correction headroom separate. A decrease in cross-entropy need not decrease the squared Bayes gap, and neither identity orders accuracy gains of fitted correctors. In Figure~\ref{fig:synthetic}, $\PH_{\mathrm{true}}=\E\|d\|^2$ is unhalved Brier headroom; its association with accuracy gain is an empirical result.

\paragraph{Feature-space reduction.}
Consider a residual feature correction $h\mapsto h+P_Sa(X)$ with $a(X)\in\R^r$, an $X$-measurable injection $P_S\in\R^{d\times r}$, and head Jacobian $J=\partial z/\partial h$. For a twice continuously differentiable convex canonical loss, form the damped Gauss--Newton surrogate
\begin{equation}
\label{eq:localq}
\begin{split}
q_i(a)&=\tfrac12a^\top B(X_i)a+g_i^\top a,\qquad
g_i=P_S^\top J_i^\top\bigl(m(z_i)-t(y_i)\bigr),\\
B(X)&=P_S^\top J^\top\nabla^2 A(z)JP_S+\kappa R(X),
\qquad \kappa>0,\quad R(X)\succ0.
\end{split}
\end{equation}
Here $R(X)$ is symmetric and $X$-measurable, so $B(X)$ is symmetric positive definite and independent of $Y$ given $X$. Label independence of the score Hessian follows from the canonical representation, not just from the CE and squared-error examples. The surrogate omits second derivatives of a nonlinear head and adds damping; it is not the exact nonlinear prediction risk.

\begin{proposition}[Feature-space reduction]
\label{prop:feature}
Under the conditions of \eqref{eq:localq}, assume the conditional moments below exist and $\E[g_i^\top B(X_i)^{-1}g_i]<\infty$. Then
\begin{equation}
\label{eq:gbar}
\bar g(X):=\E[g_i\mid X]=P_S^\top J^\top d(X),
\end{equation}
and the unique minimizer, up to null sets, of the expected surrogate over measurable corrections of finite $B$-energy is
\begin{equation}
\label{eq:aB}
a_B(X)=-B(X)^{-1}P_S^\top J^\top\bigl(m(z)-\pi(X)\bigr).
\end{equation}
Strict surrogate headroom requires $P_S^\top J^\top d(X)\ne0$ on a set of positive probability. The identity in \eqref{eq:gbar} is exact; the approximation is the surrogate in \eqref{eq:localq}.
\end{proposition}

\begin{lemma}[Completion of squares]
If $B$ is symmetric positive definite and $q(a)=\frac12a^\top Ba+g^\top a$, its unique minimizer is $a^\star=-B^{-1}g$, and
\begin{equation}
q(a)=\tfrac12\|a-a^\star\|_B^2-\tfrac12g^\top B^{-1}g,
\qquad \|v\|_B^2=v^\top Bv.
\end{equation}
\end{lemma}
\begin{proof}
Expanding $\frac12(a+B^{-1}g)^\top B(a+B^{-1}g)$ gives $q(a)+\frac12g^\top B^{-1}g$. Positive definiteness gives uniqueness.
\end{proof}

\begin{proof}[Proof of Proposition~\ref{prop:feature}]
Conditioning on $X$ fixes $P_S,J,z,B$ and replaces $t(Y)$ by $\pi(X)$, proving \eqref{eq:gbar}. For any admissible measurable $a$,
\[
\E[q_i(a)\mid X]=\tfrac12a^\top B(X)a+\bar g(X)^\top a.
\]
Completion of squares therefore gives
\[
\E[q_i(a)-q_i(a_B)\mid X]=\tfrac12\|a-a_B\|_{B(X)}^2.
\]
The moment assumption and conditional Jensen ensure that $a_B$ has finite $B$-energy; Cauchy--Schwarz makes the linear terms integrable for every correction in this class. Integrating the nonnegative difference proves optimality and uniqueness up to null sets.
\end{proof}

\subsection{Output-space dominance}
\label{app:proof-output}
\begin{proposition}[Output-space dominance]
\label{prop:output}
Let $h(X)$ and the head $f_\psi$ be measurable. Let $\mathcal A$ contain measurable residual feature corrections $h\mapsto h+P_Sa(X)$, and let $\mathcal D$ contain all measurable score edits $z\mapsto z+\omega(X)$. Every $a\in\mathcal A$ induces
\[
\omega_a(X)=f_\psi(h(X)+P_Sa(X))-f_\psi(h(X)).
\]
Thus $\inf_{\omega\in\mathcal D}R(\omega)\le\inf_{a\in\mathcal A}R(a)$ for a common prediction risk whenever these risks are defined. For an affine head $f_\psi(h)=Gh+c$, $\omega_a=GP_Sa$. Under Theorem~\ref{thm:reduction}, any finite Bayes-optimal edit satisfies $m(z+\omega^\star)=\pi(X)$, and any edit attaining this target is optimal.
\end{proposition}

\begin{proof}
The constructed edit is measurable and gives exactly the same scores as the feature correction. Inclusion of predictor classes proves the risk inequality; this comparison does not include different parameter penalties or computational constraints. For a canonical convex loss, the gradient of the conditional edit risk is $m(z+\omega)-\pi(X)$, so the optimum follows from Theorem~\ref{thm:reduction}. For squared error, $\omega^\star=\pi-z$; for categorical CE, boundary targets may only be approached in the limit.

If $f_\psi$ is differentiable at $h$, its local expansion is $f_\psi(h+P_Sa)=f_\psi(h)+JP_Sa+o(\|a\|)$ for fixed $P_S$. The remainder is $O(\|a\|^2)$ if the Jacobian is locally Lipschitz, for example under a locally bounded continuous second derivative. Exact measurable containment above does not require either local expansion.
\end{proof}

\subsection{Headroom and ambiguity}
\label{app:ambiguity}
Under the moment assumptions of Proposition~\ref{prop:feature}, completion of squares yields
\begin{equation}
\E[q_i(a(X))\mid X]=\tfrac12\|a(X)-a_B(X)\|_{B(X)}^2+\E[q_i(a_B(X))\mid X].
\end{equation}
If $m(z)=\pi(X)$ almost surely, then $\bar g=0$ and $a_B=0$: every nonzero correction on a positive-probability set strictly increases the expected damped surrogate. This strictness concerns $q$, not necessarily the undamped nonlinear prediction risk. Conversely, optimal surrogate improvement is
\begin{equation}
\E[q_i(0)]-\E[q_i(a_B)]=\tfrac12\E[\bar g^\top B^{-1}\bar g],
\end{equation}
strictly positive exactly when $\bar g\ne0$ on a set of positive probability. It does not guarantee that taking the full step $a_B$ decreases the actual nonlinear loss.

Write $g=\bar g+(g-\bar g)$. Since $B^{-1}\bar g$ is $X$-measurable and $\E[g-\bar g\mid X]=0$, the cross term vanishes, giving the oracle-energy decomposition
\begin{equation}
\E[g^\top B^{-1}g\mid X]=\bar g^\top B^{-1}\bar g+\E[(g-\bar g)^\top B^{-1}(g-\bar g)\mid X].
\end{equation}
For square-integrable $t(Y)$, let $X^{\mathrm{full}}$ include the full observation and the masking context, so that $\sigma(X)\subseteq\sigma(X^{\mathrm{full}})$. The law of total covariance gives
\begin{equation}
\begin{aligned}
\Cov(t(Y)\mid X)
&=\E[\Cov(t(Y)\mid X^{\mathrm{full}})\mid X]\\
&\quad+\Cov\bigl(\E[t(Y)\mid X^{\mathrm{full}}]\mid X\bigr).
\end{aligned}
\end{equation}
The first term is ambiguity remaining with that full information; the second comes from full states that the poorer observation cannot distinguish. For random informative masks, including the mask in $X^{\mathrm{full}}$ is necessary for the asserted nesting.

\subsection{Cross-mask inheritance and transfer}
\label{app:inherit}
Throughout, $\mathcal G_S=\sigma(X_S)\subseteq\mathcal G_T=\sigma(X_T)$ under one joint law, and $t(Y)$ is integrable. Fixed nested raw modality sets satisfy this condition. For random masks, the richer observation must also contain the poorer mask context, or the claim must be applied conditionally on a fixed context. Set inclusion $S\subseteq T$ alone does not establish sigma-field inclusion when the mask carries information about $Y$. Separately computed frozen representations also need not be nested.

Write $\pi_S=\E[t(Y)\mid X_S]$, $\pi_T=\E[t(Y)\mid X_T]$, $d_S=m_S-\pi_S$, $d_T=m_T-\pi_T$, and $b=\E[d_T\mid X_S]$. The results below give an identity, contraction bounds, a population recovery condition and a comparison of retrieval targets.

\begin{lemma}[Inheritance identity]
\label{lem:inherit}
Under the nesting above, $\pi_S=\E[\pi_T\mid X_S]$ and
\[
d_S=\bigl(m_S-\E[m_T\mid X_S]\bigr)+\E[d_T\mid X_S].
\]
\end{lemma}
\begin{proof}
The tower property gives $\E[\E[t(Y)\mid X_T]\mid X_S]=\pi_S$. Hence $\E[d_T\mid X_S]=\E[m_T\mid X_S]-\pi_S$; rearrangement proves \eqref{eq:inherit}.
\end{proof}

\begin{lemma}[The inherited term contracts]
\label{lem:contract}
If $d_T\in L^2$, then $\E\|b\|^2\le\E\|d_T\|^2$. For categorical probability vectors and $\tau_S=\E[m_T\mid X_S]$,
\begin{equation}
\label{eq:kl-contract}
\mathrm{KL}(\pi_S\|\tau_S)\le\E[\mathrm{KL}(\pi_T\|m_T)\mid X_S]
\end{equation}
almost surely, and the same inequality holds after taking expectations on both sides. KL is interpreted in the extended nonnegative sense at boundary probabilities.
\end{lemma}
\begin{proof}
Conditional Jensen gives $\|\E[d_T\mid X_S]\|^2\le\E[\|d_T\|^2\mid X_S]$; integration proves the first claim. Joint convexity of KL applied to the pair $(\pi_T,m_T)$, whose conditional mean is $(\pi_S,\tau_S)$, proves \eqref{eq:kl-contract}. The same argument applies coordinatewise to Bernoulli KL for multi-label outputs, followed by the required sum or mean. These bounds measure conditional prediction error, which is stronger than calibration aggregated over equal predicted probabilities.
\end{proof}

\begin{proposition}[Recovery condition, and non-identifiability]
\label{prop:identif}
The population transfer target $\tau_S=\E[m_T\mid X_S]$ equals $\pi_S$ almost surely if and only if $b=0$ almost surely. For a fixed frozen predictor and unlabeled paired law, $\pi_S$ is not identifiable over unrestricted conditional label laws with at least two distinct attainable label statistics.
\end{proposition}
\begin{proof}
Inheritance gives $\tau_S=\pi_S+b$, proving the equivalence. For non-identifiability, keep the paired feature law and predictor fixed. Choosing $Y=y_0$ almost surely in one law and $Y=y_1$ almost surely in another, with $t(y_0)\ne t(y_1)$, leaves every unlabeled observation unchanged but changes $\pi_S$. A condition linking predictions to labels is therefore required. This is a population statement: a finite-neighbor average need not equal $\tau_S$ even when $b=0$.
\end{proof}

\label{app:transfer}
\begin{proposition}[Transfer guarantees]
\label{prop:transfer}
Let $p_\beta(q)=(1-\beta)m_S+\beta q$, with $\beta\in[0,1]$. In both parts, the query label has its deployment conditional law independently of the retrieval sample given $X_S$.
\begin{enumerate}[leftmargin=1.8em,itemsep=2pt,topsep=2pt,label=\emph{(\roman*)}]
\item \emph{Finite-sample dominance.} For a source pool of size $n$, let $\mathcal F$ fix the query, base prediction, selected richer-view neighbor features and nonnegative weights summing to one. Suppose $\E[t(Y_j)\mid\mathcal F]=\pi_T(X_T^{(j)})$ for every selected neighbor, and $m_T=\pi_T$. For the same fixed $\beta$, any admissible $1\le k\le n$, and convex conditional prediction risk $R_{X_S}$ (including ideal CE and Brier),
\[
\E R_{X_S}\bigl(p_\beta(\widehat\pi^{(T)})\bigr)
\le\E R_{X_S}\bigl(p_\beta(\widehat\pi^{(Y)})\bigr).
\]
The expectation is over queries and retrieval samples.
\item \emph{Oracle blend and estimation error.} For squared risk, assume $t(Y)$, $d_S$ and $\varepsilon_n=\widehat\pi_n-\pi_S$ are square-integrable. Set
\[
D=\E\|d_S\|^2,\qquad V_n=\E\|\varepsilon_n\|^2,\qquad C_n=\E\langle d_S,\varepsilon_n\rangle.
\]
The unhalved excess risk of a scalar blend is
\begin{equation}
\label{eq:blend-moments}
F_n(\beta)=(1-\beta)^2D+2\beta(1-\beta)C_n+\beta^2V_n.
\end{equation}
For $A_n=D+V_n-2C_n>0$, its minimizer is
\begin{equation}
\label{eq:betastar}
\beta_n^\star=\left[\frac{D-C_n}{D+V_n-2C_n}\right]_{[0,1]}.
\end{equation}
If $A_n=0$, every $\beta\in[0,1]$ attains $F_n(\beta)=D$. If the unconstrained minimizer is in $[0,1]$, the minimum is $(DV_n-C_n^2)/A_n$; otherwise it is the appropriate endpoint value $D$ or $V_n$.

Let $n_U$ and $n_L$ be the unlabeled and labeled pool sizes. If $\widehat\pi^{(T)}_{n_U}\to\tau_S=\pi_S+b$ in $L^2$ as $n_U\to\infty$, then
\begin{equation}
\label{eq:transfer-floor}
\min_{\beta\in[0,1]}F^{(T)}_{n_U}(\beta)\longrightarrow
\min_{\beta\in[0,1]}\E\|(1-\beta)d_S+\beta b\|^2.
\end{equation}
For hard-label retrieval, $\min_\beta F^{(Y)}_{n_L}=DV_{n_L}/(D+V_{n_L})$ when $C_{n_L}=0$ and $D+V_{n_L}>0$. More generally, if $D>0$, $V_{n_L}>0$ tends to zero and $C_{n_L}^2=o(V_{n_L})$, then $\min_\beta F^{(Y)}_{n_L}=V_{n_L}(1+o(1))$. If $D=0$, choosing $\beta=0$ already gives zero excess risk.
\end{enumerate}
\end{proposition}

\begin{proof}
\emph{Part (i).} The conditional-mean assumption and the $\mathcal F$-measurable weights give
\[
\E[\widehat\pi^{(Y)}\mid\mathcal F]
=\sum_{j\in\mathcal N_k}w_j\pi_T(X_T^{(j)})=\widehat\pi^{(T)}.
\]
The map $q\mapsto R_{X_S}((1-\beta)m_S+\beta q)$ is convex for $\beta\in[0,1]$. Conditional Jensen therefore gives
\[
R_{X_S}\bigl(p_\beta(\widehat\pi^{(T)})\bigr)
\le\E\left[R_{X_S}\bigl(p_\beta(\widehat\pi^{(Y)})\bigr)\mid\mathcal F\right].
\]
Taking expectations proves the claim. Nonnegative CE may take the value $+\infty$ when a hard-label target has zeros, in which case Jensen is understood in the extended sense. Independence of neighbor labels is not needed for this step; preservation of their conditional means is. For i.i.d.\ source samples, a query independent of the pool, and label-independent fixed feature maps and neighbor selection, that condition holds. It also holds for $\mathcal F$-measurable blends if conditioning on their construction preserves the stated conditional means.

Choosing different blends, temperatures or neighbors by inspecting retrieval labels is not covered by this comparison. Even a separate fit split can select settings using predictions that depend on those labels, so disjointness alone does not justify conditioning on the selected settings. Tempering $m_T=\pi_T$ generally changes the conditional target and removes the hypothesis. The proposition compares the two fixed candidates in convex prediction risk, before fitting separate gates, and makes no claim about their accuracy ordering.

\emph{Part (ii).} The residual is $p_\beta(\widehat\pi_n)-\pi_S=(1-\beta)d_S+\beta\varepsilon_n$. Expanding its squared norm gives \eqref{eq:blend-moments}. Equivalently,
\[
F_n(\beta)=D-2\beta(D-C_n)+\beta^2A_n,
\qquad A_n=\E\|d_S-\varepsilon_n\|^2\ge0.
\]
For $A_n>0$, completion of squares and projection onto $[0,1]$ prove \eqref{eq:betastar}. Substitution gives the interior minimum. If $A_n=0$, $d_S=\varepsilon_n$ almost surely, so the residual and risk are constant in $\beta$. The formula $DV_n/(D+V_n)$ is the special case $C_n=0$. Conditional unbiasedness $\E[\varepsilon_n\mid X_S]=0$ suffices for $C_n=0$; unconditional mean zero alone does not.

For the transfer limit, put $e_n=\widehat\pi^{(T)}_{n_U}-\tau_S$ and $F_\infty(\beta)=\E\|(1-\beta)d_S+\beta b\|^2$. Cauchy--Schwarz gives
\[
\sup_{\beta\in[0,1]}|F^{(T)}_{n_U}(\beta)-F_\infty(\beta)|
\le 2(\|d_S\|_{L^2}+\|b\|_{L^2})\|e_n\|_{L^2}+\|e_n\|_{L^2}^2\longrightarrow0.
\]
Uniform convergence implies convergence of minima, proving \eqref{eq:transfer-floor}. For the hard-label asymptotic claim, $A_n\to D>0$ and the unconstrained minimizer is eventually positive. If it is at most one, the minimum divided by $V_n$ is $(D-C_n^2/V_n)/A_n\to1$; if it exceeds one, the constrained minimum is exactly $V_n$. This covers both cases.
\end{proof}

\paragraph{Neighborhood bias and variance.}
The moment formula includes smoothing bias in both retrieval modes. To make this explicit, condition on the query and the selected poorer-view features, with weights determined by those features. Assume the selected source responses are conditionally independent given them. Write $\Sigma_Y(x)=\Cov(t(Y)\mid X_S=x)$. Then
\begin{align}
a_n^{(Y)}&=\sum_jw_j\pi_S(X_S^{(j)})-\pi_S(X_S),\\
\Cov(\widehat\pi^{(Y)}\mid X_S,\{X_S^{(j)}\})&=\sum_jw_j^2\Sigma_Y(X_S^{(j)}),\\
V_n^{(Y)}&=\E\|a_n^{(Y)}\|^2+\E\sum_jw_j^2\operatorname{tr}\Sigma_Y(X_S^{(j)}),
\qquad C_n^{(Y)}=\E\langle d_S,a_n^{(Y)}\rangle.
\end{align}
For transfer, replace the conditional mean $\pi_S$ by $\tau_S=\pi_S+b$ and $\Sigma_Y$ by $\Sigma_T(x)=\Cov(m_T(X_T)\mid X_S=x)$. Its conditional bias relative to $\pi_S$ is $b(X_S)+\sum_jw_j\tau_S(X_S^{(j)})-\tau_S(X_S)$, not just $b(X_S)$ at a finite bandwidth.

When $m_T=\pi_T$, total covariance gives
\[
\Sigma_Y(X_S)=\E[\Cov(t(Y)\mid X_T)\mid X_S]+\Cov(\pi_T\mid X_S),
\]
so $\Sigma_T(X_S)\preceq\Sigma_Y(X_S)$. This is a matrix inequality; their traces give the scalar noise contributions to squared risk. Under the fixed weighted comparison above, the two modes share the same smoothing bias and the transfer mode has no greater conditional covariance. For uniform weights in a neighborhood of radius $h$ around $x$, continuity of the covariance gives $\Sigma(x)/k+o(1/k)$ as $h\to0$, rather than an exact finite-pool formula. A Lipschitz conditional mean gives smoothing bias $O(h)$. In an integrated regime with bounded constants and nonvanishing integrated noise of order $1/k$, $kh^2\to0$ makes its squared contribution $o(1/k)$ and implies $C_n^2=o(V_n)$. These are additional regularity and sampling conditions; they do not follow from the existence of a finite retrieval pool. Weighted neighbors instead contribute through $\sum_jw_j^2$.

More generally, if an estimator has the exact decomposition $\varepsilon=b_n(X_S)+\xi$ with $\E[\xi\mid X_S]=0$, its blend denominator is
\[
D+V-2C=\E\|d_S-b_n\|^2+\E\|\xi\|^2,
\qquad C=\E\langle d_S,b_n\rangle.
\]
The integrated noise $\E\|\xi\|^2$ must not be replaced by an unaveraged, query-dependent covariance. In the special unbiased case $b_n=0$, lower integrated noise lowers the oracle risk $DV/(D+V)$ for $D>0$.

\begin{remark}[Two limits, and what they do not say]
\label{rem:floors}
Under the $L^2$ consistency assumption in Proposition~\ref{prop:transfer}, the transfer target's MSE tends to $\E\|b\|^2$, while the oracle blend's excess risk tends to
\[
\min_{\beta\in[0,1]}\E\|(1-\beta)d_S+\beta b\|^2
\le\min\{D,\E\|b\|^2\}.
\]
The target limit follows from $L^2$ convergence, and the inequality follows by choosing either endpoint. Consistency requires estimation error to vanish; increasing the pool with a fixed $k$ need not suffice. These statements concern an oracle scalar blend and retrieval labels only. GUARD still uses labels for fitting and calibration, selects $\widehat\beta$ on a finite grid, and applies a learned gate. No consistency or risk ordering for that fitted policy follows from the oracle formula.
\end{remark}

\subsection{Finite-sample harm control}
\label{app:noharm}
\paragraph{Conditioning and the loss family.}
Let $\mathcal H$ contain all fitting data, fixed randomization and choices made before calibration: the retrieval pool and rule, reference $p_{\base}$, candidate $p_{\corr}$, feature transforms, temperature, blend, scorer, prediction-loss convention and tolerance $\delta$. Conditional on $\mathcal H$, assume the $n$ calibration samples and the next deployment sample are exchangeable. Disjoint data roles alone are not a substitute for this assumption. The fitted objects define a common measurable function for each sample; any query-dependent retrieval weighting must follow that same rule at calibration and deployment.

Take finite scores and thresholds in $\overline{\R}$, with $A_{-\infty}=1$ and $A_{+\infty}=0$. The functions $L_i(\lambda)$ in \eqref{eq:policy} are exchangeable conditional on $\mathcal H$, bounded by one, non-increasing and right-continuous, because the gate uses $r_\theta>\lambda$. At $+\infty$ the loss is zero. These are the conditions of conformal risk control~\citep{angelopoulos2024crc}. If $n=0$ or the corrected empirical criterion is infeasible, set $\widehat\lambda=+\infty$.

\begin{proof}[Proof of Proposition~\ref{thm:noharm}]
For $n\ge1$, define the symmetric auxiliary threshold
\[
\lambda^\circ=\inf\left\{\lambda:\frac1{n+1}\sum_{i=1}^{n+1}L_i(\lambda)\le\alpha\right\}.
\]
The finite step functions and the no-intervention endpoint ensure feasibility at the infimum. Since $L_{n+1}\le1$, every feasible CRC threshold is feasible here, so $\widehat\lambda\ge\lambda^\circ$, also under the empty-set convention. Monotonicity and conditional exchangeability give
\[
\E[L_{n+1}(\widehat\lambda)\mid\mathcal H]
\le\E[L_{n+1}(\lambda^\circ)\mid\mathcal H]
=\E\left[\frac1{n+1}\sum_iL_i(\lambda^\circ)\,\middle|\,\mathcal H\right]\le\alpha.
\]
For $n=0$ the loss is zero by construction. Integrating over $\mathcal H$ proves \eqref{eq:noharm}.
\end{proof}

The probability is marginal over calibration and deployment, and is relative to the declared $p_{\base}$, loss and tolerance. It bounds joint harmful intervention, not harm conditional on applying, mean loss increase, or a benchmark metric such as AUROC. An unbounded ideal CE loss poses no problem because the calibrated loss is its bounded harm indicator. At $\alpha<1/(n+1)$, no threshold satisfies the corrected criterion; the implementation also chooses no intervention at equality, a conservative convention.

\paragraph{Tightening and what must remain fixed.}
For any measurable post-calibration choice $\lambda^\star\ge\widehat\lambda$, $L_{\lambda^\star}\le L_{\widehat\lambda}$ for every calibration draw and deployment point. Thus tightening preserves the bound even if the chosen threshold depends on calibration. It must retain the same score, candidate, reference, loss and tolerance; changing any of these changes the loss family and is not justified by threshold monotonicity. GUARD uses $D_{\mathrm{fit}}$ for its utility-based tightening.

\paragraph{Mask-conditional calibration.}
Use a partition by masks fixed before calibration. For a group $s$, apply \eqref{eq:crc} to its $n_s$ calibration observations and use no intervention when $n_s=0$.
\begin{corollary}[Mask-conditional harm control]
\label{cor:mask}
Suppose, conditional on $\mathcal H$, $S_{\mathrm{test}}=s$ and the group count $n_s$, the group calibration observations and the deployment observation are exchangeable under the group law. Then the group threshold satisfies
\[
\Prob(\Delta_Y(X)>\delta\ \wedge\ \apply\mid S_{\mathrm{test}}=s)\le\alpha.
\]
\end{corollary}
\begin{proof}
For each nonempty group and each count, apply the preceding proof under the stated conditional law; empty or infeasible groups have zero intervention loss. Average over the counts and fitting information conditional on $S_{\mathrm{test}}=s$.
\end{proof}

\paragraph{Cost of a poor scorer.}
A poor score can reduce utility without invalidating this bound. A constant score offers only blanket intervention or no intervention. Blanket intervention is admitted only if its calibration harm satisfies $(n\widehat R_n+1)/(n+1)\le\alpha$, not merely $\widehat R_n\le\alpha$. The agreement-only examples on NinaPro and AVE illustrate these two policies. Neither policy certifies the outcome of an individual query.

\paragraph{Learn-then-Test comparator.}
Learn-then-Test~\citep{angelopoulos2021ltt} instead provides a high-probability risk statement over the calibration draw. Its Hoeffding--Bentkus tests require i.i.d.\ calibration losses (or a separately justified valid concentration bound), a stronger condition than arbitrary exchangeability. Our comparator fixes its candidate threshold grid and order using $D_{\mathrm{fit}}$, tests $H_\lambda:R(\lambda)>\alpha$ in sequence at level $\zeta=0.05$, and stops at the first non-rejection. The first true null, if reached, is rejected with probability at most $\zeta$, which controls erroneous certification along the sequence. The risk being tested is the same joint harm. Tightening keeps this statement too: on the event that the selected threshold has risk at most $\alpha$, any larger threshold does as well.

\paragraph{An ambiguity floor on conditional harm.}
Conformal risk control bounds joint harm. For corrections large enough to cause harm, no gate can bring the harm rate below the label ambiguity of those samples.

\begin{proposition}[Harm floor for binary labels]
\label{prop:floor}
Consider binary labels under cross-entropy, a deployment pair $(X,Y)$ independent of the fitting and calibration data, a policy $\apply$ measurable in $X$ given those data, and the event $E=\{\max_y\Delta_y(X)>\delta\}$, on which the correction raises the loss beyond the tolerance for at least one label. If $\Prob(\apply,E)>0$, then
\begin{equation}
\label{eq:floor}
\Prob\bigl(\Delta_Y(X)>\delta\mid\apply,E\bigr)\ \ge\ \E\bigl[\min\{\pi_0(X),\pi_1(X)\}\mid\apply,E\bigr].
\end{equation}
\end{proposition}
\begin{proof}
On $E$ the two predictions differ. Both sum to one, so exactly one label $y^\star(X)$ loses probability and only that label can have $\Delta_y(X)>0$. Hence $\Prob(\Delta_Y>\delta\mid X)=\pi_{y^\star(X)}(X)\ge\min\{\pi_0(X),\pi_1(X)\}$ on $E$. Conditional on the fitting and calibration data, $\apply$ and $E$ are functions of $X$ and $Y$ keeps its conditional law given $X$, so averaging over $\{\apply,E\}$ gives \eqref{eq:floor}.
\end{proof}

The right side is the Bayes error on the samples in $E$. The reported conditional harm also counts applied samples outside $E$, which cannot be harmful: $\Prob(\Delta_Y>\delta\mid\apply)=\Prob(E\mid\apply)\,\Prob(\Delta_Y>\delta\mid\apply,E)$, and \eqref{eq:floor} bounds only the second factor. Bayes error is unknown on real data, so the floor limits what selection can achieve and does not estimate an observed rate. It also does not contradict useful correction. An overconfident $p_1=0.9$ corrected to the Bayes value $\pi_1=0.7$ raises the loss by $0.25$ for the label that occurs with probability $0.7$, yet lowers expected cross-entropy by $0.15$: an exact correction with conditional harm $0.7$. With more than two classes, some label always loses probability, which gives only the weaker bound $\min_y\pi_y(X)$.

\begin{figure}[!htbp]
\centering
\includegraphics[width=\linewidth]{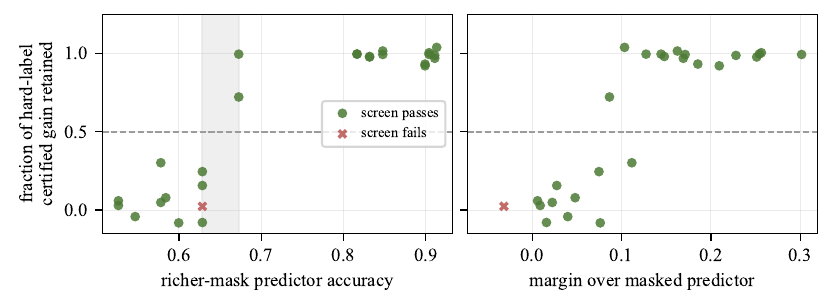}
\caption{\textbf{Retained cross-mask gain against two screening statistics.} Twenty-eight dataset--condition--seed cells from binary DrugBAN and five-class OPPORTUNITY have hard-label gated gain at least $0.02$. \emph{Left:} richer-mask accuracy separates the observed modes, with no cell between $0.628$ and $0.673$. \emph{Right:} the margin over the masked predictor overlaps between $0.086$ and $0.111$. The separation persists for headroom cuts $0.01$--$0.03$ and fails at $0.005$. This is a descriptive comparison, not a universal accuracy threshold or a test of conditional correctness.}
\label{fig:screen}
\end{figure}

\subsection{Selecting the blend weight}
\label{app:erm}
The action score $r_\theta$ is a logistic model fitted on $D_{\mathrm{fit}}$ against $Z=\mathbf1\{\Delta_Y<0\}$. Its five features are the largest reference probability, reference entropy, largest target probability, the agreement flag $\mathbf1\{K^{-1}\sum_c|p_{\base,c}-\widehat\pi_c|<0.1\}$, and the $\ell_1$ discrepancy $\sum_c|p_{\base,c}-\widehat\pi_c|$. Probabilities used in the reference features are clipped for numerical stability. Categorical entropy is $-\sum_cp_c\log p_c$; for multi-label predictions GUARD uses $-\sum_c[p_c\log p_c+(1-p_c)\log(1-p_c)]$, the sum of marginal Bernoulli entropies.

For two categorical simplex rows, $\sum_c|p_c-q_c|\le2$, so the agreement flag is identically one when $C>20$, as on AVE and NinaPro DB5; it carries no signal there. The coordinate clipping changes this distance by at most $C\,10^{-12}$ and does not alter that conclusion for these benchmarks. This simplex argument does not apply to multi-label vectors. The feature is retained as implemented; a different scaling would require evaluating a different scorer.

The blend is selected from $41$ equally spaced values in $[0,1]$ by minimizing mean prediction loss $\bar\ell$ on $D_{\mathrm{fit}}$, before calibration. Retrieval settings are selected using the benchmark metric, except for the accuracy-based comparison in Table~\ref{tab:gates}. This fit criterion is separate from the ideal oracle analysis of Proposition~\ref{prop:transfer}; no generalization or consistency bound for the grid selector is claimed.

\paragraph{Utility-based tightening.}
Let $g_i=-\Delta_{Y_i}(X_i)$ on $D_{\mathrm{fit}}$. Among the calibrated threshold and distinct fit scores above it, evaluate
\[
G(\lambda)=\sum_{i:r_\theta(X_i)>\lambda}g_i,
\qquad s(\lambda)=\sqrt{N_\lambda}\,\widehat{\operatorname{sd}}\{g_i:r_\theta(X_i)>\lambda\},
\]
where $N_\lambda$ is the number admitted and $s=0$ when fewer than two predictions are admitted. If the best gain is positive, retain the largest candidate threshold whose gain is within $s(\lambda_{\mathrm{best}})$ of that best gain. Otherwise move above all fit scores, without dropping below the CRC threshold. This is the implemented one-standard-error heuristic on accepted gains; it is not a confidence interval for total deployment utility. Every selected threshold is at least $\widehat\lambda_{\mathrm{CRC}}$, which is the property needed for harm control.

For fixed data and score, the CRC threshold alone is non-increasing in $\alpha$. The subsequent utility search need not preserve that ordering: it changes both the best fit gain and the set of near-best thresholds as the admissible set expands. Figure~\ref{fig:apply} therefore describes an empirical intervention curve, not a monotonicity theorem for the tightened policy.

All fitted choices optimize a risk on the fit distribution $Q$. Matching deployment $Q'$ requires more than reducing sampling error on $Q$; additional fit data do not remove a distribution mismatch. Our fitting splits are drawn from the evaluated deployment condition. The harm guarantee separately requires the conditional exchangeability stated in App.~\ref{app:noharm}.

\subsection{The cross-mask accuracy screen}
\label{app:screen}
The population accuracy ordering gives a necessary condition for a conditionally correct richer predictor. For categorical labels and any poorer-view classifier $a_S(X_S)$,
\begin{equation}
\label{eq:accuracy-screen}
\E\max_c\pi_{T,c}\ \ge\ \E\max_c\pi_{S,c}\ \ge\ \Prob(a_S(X_S)=Y).
\end{equation}
The first inequality follows from conditional Jensen for the maximum and the tower identity; the second is the Bayes-optimality of $\arg\max_c\pi_{S,c}$. If $m_T=\pi_T$, its argmax attains the left side. Thus a strictly lower \emph{population} accuracy rules out conditional correctness. Passing this ordering is not sufficient: accurate class decisions need not estimate conditional probabilities correctly.

Finite-sample accuracy reversals are descriptive, not statistical rejections without an appropriate uncertainty assessment. Among the twenty-eight dataset--condition--seed cells of Figure~\ref{fig:screen} with hard-label gated gain at least $0.02$, the ordering flags two, both retaining essentially none of that gain. Ten of the twenty-six passing cells also retain none. Here, richer-mask accuracy separates outcomes more clearly than its margin over the poorer predictor.

Absolute accuracy depends on class cardinality and balance: uniform guessing scores $1/2$ on binary DrugBAN and $1/5$ on five-class OPPORTUNITY, while majority-class baselines depend on label frequencies. The observed separation does not transfer directly to $29$-way AVE or many-class NinaPro.

Retained fractions here exceed $0.72$ or fall below $0.31$. Lemma~\ref{lem:contract} bounds inherited error by richer-view error in expected squared norm and KL. These population bounds do not imply continuity of accuracy or gated gain, which also depend on finite retrieval, tuning and discrete decisions.

\end{document}